\documentclass[a4paper,twoside]{book}
\usepackage[a4paper,includeheadfoot,margin=2.54cm]{geometry}

\usepackage[T1]{fontenc}
\usepackage[utf8]{inputenc}
\usepackage{textcomp}
\usepackage{graphicx}
\usepackage[ruled]{algorithm2e}
\usepackage{lipsum}
\usepackage[normalem]{ulem}
\usepackage{booktabs}
\usepackage[small,bf]{caption}
\usepackage[round]{natbib}
\usepackage{example}
\usepackage{times}
\usepackage{tipa}
\usepackage{todonotes}
\usepackage{amsmath}
\usepackage{lscape}

\usepackage[hidelinks, unicode=true, pdftitle={Final Report of the Frederick Jelinek Memorial Summer Workshop on Neural Polysynthetic Language Modelling},pdfsubject={Neural Polysynthetic Language Modelling}, pdfauthor={Lane Schwartz, Francis Tyers, Lori Levin, Christo Kirov, Patrick Littell, Chi-kiu (Jackie) Lo, Emily Prud’hommeaux, Hyunji (Hayley) Park, Kenneth Steimel, Rebecca Knowles, Jeffrey Micher, Lonny Strunk, Han Liu, Coleman Haley, Katherine J. Zhang, Robbie Jimmerson, Vasilisa Andriyanets, Aldrian Obaja Muis, Naoki Otani, Jong Hyuk (Jay) Park, and Zhisong Zhang},pdfkeywords={language model, language modelling, neural language modelling, polysynthetic language, endangered languages, neural polysynthetic language modelling, machine translation, finite-state morphology, predictive text, BPE, Chukchi, St. Lawrence Island Yupik, Central Alaskan Yup’ik, Inuktitut, Crow, Guaraní, tensor product representation}]{hyperref}

\usepackage[capitalize,noabbrev]{cleveref}

\usepackage{amsfonts}
\usepackage{xcolor}

\newcounter{InterlinearGloss}
\newcommand{\ilg}[1]{(\refstepcounter{InterlinearGloss}\arabic{InterlinearGloss}\label{#1})}
\crefformat{InterlinearGloss}{example (#1)}
\crefname{InterlinearGloss}{example}{examples}
\Crefname{InterlinearGloss}{Example}{Examples}

\newcommand{\CKT}{\texttt{ckt}}
\newcommand{\ckt}{Chukchi}

\newcommand{\CRO}{\texttt{cro}}
\newcommand{\cro}{Crow}

\newcommand{\ENG}{\texttt{eng}}
\newcommand{\Eng}{English}

\newcommand{\Es}{Yupik} 

\newcommand{\ESS}{\texttt{ess}}
\newcommand{\Ess}{St.~Lawrence Island Yupik}
\newcommand{\ess}{St.~Lawrence Island Yupik}

\newcommand{\ESU}{\texttt{esu}}
\newcommand{\Esu}{Central Alaskan Yup'ik}
\newcommand{\esu}{Central Alaskan Yup'ik}

\newcommand{\GRN}{\texttt{grn}}
\newcommand{\Grn}{Guaran\'{i}} 

\newcommand{\IKU}{\texttt{iku}}
\newcommand{\Iku}{Inuktitut}

\newcommand{\SPA}{\texttt{spa}}
\newcommand{\Spa}{Spanish}

\newcommand{\ENU}{\ENG\textsubscript{\ESU}}
\newcommand{\ENS}{\ENG\textsubscript{\ESS}}

\DeclareUnicodeCharacter{1EF9}{\~{y}}
\DeclareUnicodeCharacter{2713}{$\checkmark$}
\DeclareUnicodeCharacter{02BC}{'}
\DeclareUnicodeCharacter{263}{$\gamma$} 
\DeclareUnicodeCharacter{259}{ă} 
\newcommand{\ipa}{}

\bibdata{report}

\title{Neural Polysynthetic Language Modelling}
\begin{document}

\frontmatter

\begin{titlepage}
\centering
\vspace*{3cm}
{\Huge \bf Final Report \\}
\vspace{1cm}
{\Large \bf of the \\}
\vspace{1cm}
{\Large \bf Frederick Jelinek Memorial Summer Workshop \\}
\vspace{1cm}
{\Large \bf on \\}
\vspace{1cm}
{\Huge \bf Neural Polysynthetic \\}
\vspace{1cm}
{\Huge \bf Language Modelling \\}
\vspace{3cm}

\end{titlepage}

\begin{center}
\vspace*{85mm}
    {\Large \bf May 2020}
\end{center}
\clearpage

\begin{center}
\large
\vspace*{85mm}
Sixth Frederick Jelinek Memorial Summer Workshop \\ \ \\ 24 June -- 02 August 2019 \\ \ \\ École de Technologie Supérieure \\ Montréal, Québec, Canada
\end{center}
\clearpage


\chapter{Acknowledgements}

The work described herein was performed by the Neural Polysynthetic Language Modelling team at the Sixth Frederick Jelinek Memorial Summer Workshop, which was organized and sponsored by Johns Hopkins University with unrestricted gifts from Amazon, Facebook, Google, and Microsoft.
This work utilizes resources supported by the National Science Foundation’s Major Research Instrumentation program, grant \#1725729, as well as the University of Illinois at Urbana-Champaign.
This article contains output of a research project implemented as part of the Basic Research Programme at the National Research University Higher School of Economics (HSE University).

This workshop took place at \emph{École de technologie supérieure} in Montréal, Québec, Canada on the traditional territory of the Kanien’kehá:ka people.
The ongoing research at our home institutions in Illinois, Indiana, Pennsylvania, Maryland, Massachusetts, New York, Colorado, Washington, and Ontario takes place on the traditional territories of numerous indigenous peoples.
Our work at the University of Illinois takes place on the lands of the Peoria, Kaskaskia, Piankashaw, Wea, Miami, Mascoutin, Odawa, Sauk, Mesquaki, Kickapoo, Potawatomi, Ojibwe, and Chickasaw peoples.
Our work at Indiana University Bloomington takes place on the lands of the Miami, Lenni Lenape, Potawatomi, and Shawnee peoples.
Our work at Carnegie Mellon University takes place on the lands of the Lenni Lenape, Shawnee, and Haudenosaunee Nations.
Our work at NRC Canada in Ottawa takes place on the traditional and unceded territory of the Algonquin Nation.
Our work at Rochester Institute of Technology takes place on Onödawa’ga:’ land.
Our work at the University of Colorado Boulder takes place on the traditional lands of the Ute, Cheyenne, and Arapaho peoples.
Our work at the University of Washington takes place on the traditional lands of the Suquamish, Tulalip and Muckleshoot nations.
Our work at Boston College takes place on the traditional lands of the Mashpee Wampanoag, Aquinnah Wampanoag, Nipmuc, and Massachusett tribal nations.
Our work at Johns Hopkins University takes place on the traditional lands of the Piscataway Tribe.
Our fieldwork in Alaska takes place on the lands of St. Lawrence Island Yupik and Central Alaskan Yup'ik peoples.
We acknowledge these and all of the indigenous peoples whose lands and waters we live and work upon.

In this work we are honored to work with the languages of the St. Lawrence Island Yupik, Central Alaskan Yup'ik, Inuit, Chukchi, Crow, and Guaraní peoples.
We hope and strive for our work to serve the communities whose languages we work with.
We honor and acknowledge the rich history, languages, and cultural legacies of all of these indigenous peoples.


\chapter{Abstract}



Many techniques in modern computational linguistics and natural language processing (NLP) make the assumption that approaches that work well on English and other widely used European (and sometimes Asian) languages are ``language agnostic'' -- that is that they will also work across the typologically diverse languages of the world.
In high-resource languages, especially those that are analytic rather than synthetic, a common approach is to treat morphologically-distinct variants of a common root (such as \textit{dog} and \textit{dogs}) as completely independent word types. 
Doing so relies on two main assumptions: that there exist a limited number of morphological inflections for any given root, and that most or all of those variants will appear in a large enough corpus (conditioned on assumptions about domain, etc.) so that the model can adequately learn statistics about each variant.
Approaches like stemming, lemmatization, morphological analysis, subword segmentation, or other normalization techniques are frequently used when either of those assumptions are likely to be violated, particularly in the case of synthetic languages like Czech and Russian that have more inflectional morphology than English.

Within the NLP literature, agglutinative languages like Finnish and Turkish are commonly held up as extreme examples of morphological complexity that challenge common modelling assumptions. 
Yet, when considering all of the world’s languages, Finnish and Turkish are closer to the average case in terms of synthesis.
When we consider polysynthetic languages (those at the extreme of morphological complexity), even approaches like stemming, lemmatization, or subword modelling may not suffice. 
These languages have very high numbers of \textit{hapax legomena} (words appearing only once in a corpus), underscoring the need for appropriate morphological handling of words, without which there is no hope for a model to capture enough statistical information about those words.
Moreover, many of these languages have only very small text corpora, substantially magnifying these challenges.

To this end, we examine the current state-of-the-art in language modelling, machine translation, and predictive text completion in the context of four polysynthetic languages: Guaraní, \ess, \esu, and \Iku.
%
%
We have a particular focus on Inuit-Yupik, a highly challenging family of endangered polysynthetic languages that ranges geographically from Greenland through northern Canada and Alaska to far eastern Russia. 
The languages in this family
are extraordinarily challenging from a computational perspective, with pervasive use of derivational morphemes in addition to rich sets of inflectional suffixes and phonological challenges at morpheme boundaries.
%
%
Finally, we propose a novel framework for language modelling that combines knowledge representations from finite-state morphological analyzers with Tensor Product Representations \citep{smolensky1990} in order to enable successful neural language models capable of handling the full linguistic variety of typologically variant languages.
%




\chapter{Team Members}

\section*{Team Leader}
\addcontentsline{toc}{section}{Team Leader}

\begin{itemize}
    \item Lane Schwartz    \\
          \textit{Assistant Professor}\\
          \textit{Department of Linguistics}\\
          \textit{University of Illinois at Urbana-Champaign} \\
          \texttt{\href{mailto:lanes@illinois.edu}{lanes@illinois.edu}} \\
          \ \\
          Lane Schwartz is an Assistant Professor of Computational Linguistics at the University of Illinois at Urbana-Champaign. His research centers on computational linguistics for endangered languages, with a focus on St.~Lawrence Island Yupik; this includes work in polysynthetic language modelling, cognitively-motivated unsupervised grammar induction, and machine translation. 
          He is one of the original developers of Joshua, an open source toolkit for tree-based statistical machine translation, and was a frequent contributor to Moses, the de-facto standard for phrase-based statistical machine translation.
\end{itemize}

\section*{Senior Members \& Affiliates}
\addcontentsline{toc}{section}{Senior Members \& Affiliates}

\begin{itemize}
    \item Francis Tyers \\
          \textit{Assistant Professor}\\
          \textit{Department of Linguistics}\\
          \textit{Indiana University} \\
          \texttt{\href{mailto:ftyers@iu.edu}{ftyers@iu.edu}} \\
          \ \\
          Francis Tyers is an Assistant Professor of Computational Linguistics at Indiana University Bloomington. His research is focused on 
          language technology for marginalized and indigenous languages and communities and he has worked extensively on the Uralic languages
          and the Turkic languages. In language technology his main interests are morphological modelling, using finite-state transducers 
          and neural networks, dependency syntax and parsing, and machine translation. He is part of the core team of the Universal Dependencies
          project and secretary of the Apertium project --- a free/open-source platform for machine translation.

    \clearpage
    \item Lori Levin \\
          \textit{Research Professor} \\
          \textit{Language Technologies Institute} \\
          \textit{Carnegie Mellon University} \\
          \texttt{\href{mailto:levin@andrew.cmu.edu}{levin@andrew.cmu.edu}} \\
          \ \\
          Lori Levin is a Research Professor at the Language Technologies Institute at Carnegie Mellon University.   She has 20 years experience in NLP for low-resource and endangered languages on several funded projects.  She specializes in morphosyntax, language typology, and Construction Grammar.   

    \item Christo Kirov\\
\textit{Google} \\
\texttt{\href{mailto:ckirov@gmail.com}{ckirov@gmail.com}} \\
          \ \\
          Christo Kirov is a Research Software Engineer at Google, and was previously a Postdoctoral Fellow in the Center for Language and Speech Processing at Johns Hopkins University. His research has focused on computational morphophonology, especially in cross-linguistic, low-resource settings. He is one of the founders of the UniMorph project, which provides structured morphological paradigm data and related tools for many languages.

\item Patrick Littell \\
\textit{Research Officer}\\
\textit{National Research Council of Canada} \\
\texttt{\href{mailto:patrick.littell@nrc-cnrc.gc.ca}{patrick.littell@nrc-cnrc.gc.ca}} \\
          \ \\
          Patrick Littell is a Research Officer in the Multilingual Text Processing team at the National Research Council of Canada (NRC-CNRC).  His current research involves techniques for language technology development in very low-resource languages, by combining pre-trained multilingual models and knowledge-based rules and priors.

\item Chi-kiu (Jackie) Lo \\
\textit{Research Officer}\\
\textit{National Research Council of Canada} \\
\texttt{\href{mailto:chikiu.lo@nrc-cnrc.gc.ca}{chikiu.lo@nrc-cnrc.gc.ca}} \\
          \ \\
          Chi-kiu Lo is a Research Officer in the Multilingual Text Processing team at the National Research Council of Canada (NRC-CNRC). Her research interest is multilingual natural language processing with particular focuses on semantics in machine translation (MT), its quality evaluation and estimation. She designs a unified semantic-oriented MT quality evaluation and estimation metric, YiSi, that is readily available for evaluating translation quality in any language.

\item Emily Prud'hommeaux \\
\textit{Assistant Professor} \\
\textit{Department of Computer Science} \\
\textit{Boston College} \\
\texttt{\href{mailto:prudhome@bc.edu}{prudhome@bc.edu}} \\
          \ \\
          Emily Prud'hommeaux is an Assistant Professor of Computer Science at Boston College. Her research area is natural language and speech processing in low-resource settings, with a focus on developing tools to support the revitalization of the Haudenosaunee languages and other endangered languages of North America.


\end{itemize}

\clearpage
\section*{Graduate Students}
\addcontentsline{toc}{section}{Graduate Students}

\begin{itemize}
    \item Hyunji Hayley Park\\
\textit{Department of Linguistics}\\
\textit{University of Illinois at Urbana-Champaign} \\
\texttt{\href{mailto:hpark129@illinois.edu}{hpark129@illinois.edu}} \\
          \ \\
          Hayley Park is a PhD student in Computational Linguistics at the University of Illinois at Urbana-Champaign. Her research focuses on computational linguistics and natural language processing for low-resource languages. Her recent projects include language modelling, grammar induction, morphological analysis and corpus digitization for low-resource languages.

\item Kenneth Steimel \\
\textit{Department of Linguistics}\\
\textit{Indiana University} \\
\texttt{\href{mailto:ksteimel@iu.edu}{ksteimel@iu.edu}} \\
          \ \\
          Kenneth Steimel is a PhD candidate in Computational Linguistics at the University of Indiana Bloomington. His primary research interests are data-driven tagging of morphologically complex languages, particularly Bantu languages. His current research focuses on cross-language tagging for low resource Bantu languages.
\item Rebecca Knowles \\
\textit{Center for Language and Speech Processing, Johns Hopkins University \&}\\
\textit{National Research Council of Canada} \\
\texttt{\href{mailto:Rebecca.Knowles@nrc-cnrc.gc.ca}{Rebecca.Knowles@nrc-cnrc.gc.ca}} \\
          \ \\
          Rebecca Knowles is a Research Associate at the National Research Council of Canada (NRC-CNRC). She recently completed her Ph.D. in computer science at Johns Hopkins University. Her current research focuses on machine translation and computer aided translation.

\item Jeffrey Micher \\
\textit{Army Research Laboratory \&}\\
\textit{Carnegie Mellon University} \\
\texttt{\href{mailto:jmicher@cs.cmu.edu}{jmicher@cs.cmu.edu}} \\
          \ \\
          Jeffrey Micher is a computer science researcher at Army Research Lab and a Ph.D. student at Carnegie Mellon Universtiy.  His research interests include machine translation and morphological analysis of polysynthetic languages, specifically Inuktitut.

\item Lonny Strunk \\
\textit{Department of Linguistics}\\
\textit{University of Washington} \\
\texttt{\href{mailto:lonny.strunk@gmail.com}{lonny.strunk@gmail.com}} \\
          \ \\
          Lonny Strunk is a Master's student in the computational linguistics program at the University of Washington. His research interests focus on language technology for indigenous languages. His current project is in the creation of a finite state morphological analyzer for his heritage language of Central Alaskan Yup'ik.

\item Han Liu \\
\textit{Department of Computer Science} \\
\textit{University of Colorado Boulder} \\
\texttt{\href{mailto:han.liu@colorado.edu}{han.liu@colorado.edu}} \\
          \ \\
          Han Liu is a Ph.D. student in computer science at the University of Colorado Boulder. His research interests include natural language processing, human-centered machine learning, and human-AI collaboration.

\end{itemize}

\section*{Undergraduate Students}
\addcontentsline{toc}{section}{Undergraduate Students}

\begin{itemize}
    \item Coleman Haley \\
\textit{Johns Hopkins University} \\
\texttt{\href{mailto:chaley7@jhu.edu}{chaley7@jhu.edu}} \\
          \ \\
          Coleman Haley is an undergraduate senior at Johns Hopkins University majoring in Computer Science and Cognitive Science. His research interests include neural interpretability in natural language processing, as well as NLP for morphologically and typologically diverse languages.

\item Katherine J. Zhang \\
\textit{Carnegie Mellon University} \\
\texttt{\href{mailto:kjzhang@alumni.cmu.edu}{kjzhang@alumni.cmu.edu}} \\
          \ \\
          Katherine Zhang is a member of the teaching staff at Carnegie Mellon University's Language Technologies Institute. She recently graduated from CMU with majors in Linguistics and Chinese Studies. Her research interests lie in corpus linguistics and Sino-Tibetan languages.

\end{itemize}

\section*{Graduate Student Affiliates}
\addcontentsline{toc}{section}{Graduate Student Affiliates}

\begin{itemize}
    \item Robbie Jimerson \\
\textit{Rochester Institute of Technology} \\
\texttt{\href{mailto:rcj2772@rit.edu}{rcj2772@rit.edu}} \\
          \ \\
          Robbie Jimerson is a Ph.D. candidate in Computing and Information Sciences at the Rochester Institute of Technology and a member of the Seneca Nation of Indians. His dissertation research focuses on developing robust language technologies to support the documentation and revitalization of the Seneca language and other endangered indigenous languages.

\item Vasilisa Andriyanets \\
\textit{Moscow Higher School of Economics} \\
\texttt{\href{mailto:blindedbysunshine@gmail.com}{blindedbysunshine@gmail.com}} \\
          \ \\
	 Vasilisa Adriyanets is a recently graduated Masters student in computational linguistics from Higher School of 
         Economics in Moscow, Russia. She has worked on computational approaches to processing a variety of languages, 
         and specifically on morphological analysis for Russian and Chukchi.
\end{itemize}

\section*{Remote Student Affiliates}
\addcontentsline{toc}{section}{Remote Student Affiliates}

\begin{itemize}
    
    \item Aldrian Obaja Muis, Naoki Otani, Jong Hyuk (Jay) Park, Zhisong Zhang \\
\textit{Carnegie Mellon University} \\
\texttt{\{amuis@cs, notani@cs, jp1@andrew, zhisongz@andrew\}.cmu.edu} \\
          \ \\

\end{itemize}


\tableofcontents
\addcontentsline{toc}{chapter}{Contents}

\mainmatter


\chapter{Introduction\label{ch:intro}}

Many techniques in modern computational linguistics and natural language processing (NLP) make the assumption that approaches that work well on English and other widely used European (and sometimes Asian) languages are ``language agnostic'' -- that is that they will also work across the typologically diverse languages of the world.\footnote{Emily Bender provides a thorough discussion of this problem in \url{https://thegradient.pub/the-benderrule-on-naming-the-languages-we-study-and-why-it-matters/}.}
In high-resource languages, especially those that are analytic rather than synthetic, a common approach is to treat morphologically-distinct variants of a common root (such as \textit{dog} and \textit{dogs}) as completely independent word types. 
Doing so relies on two main assumptions: that there exist a limited number of morphological inflections for any given root, and that most or all of those variants will appear in a large enough corpus (conditioned on assumptions about domain, etc.) so that the model can adequately learn statistics about each variant.
Approaches like stemming, lemmatization, morphological analysis, subword segmentation, or other normalization techniques are frequently used when either of those assumptions are likely to be violated, particularly in the case of synthetic languages like Czech and Russian that have more inflectional morphology than English.

Within the NLP literature, agglutinative languages like Finnish and Turkish are commonly held up as extreme examples of morphological complexity that challenge common modelling assumptions. 
Yet, when considering all of the world's languages, Finnish and Turkish are closer to the average case in terms of synthesis.
When we consider polysynthetic languages (those at the extreme of morphological complexity), approaches like stemming, lemmatization, or subword modelling may not suffice.
These languages have very high numbers of \textit{hapax legomena} (words appearing only once in a corpus), underscoring the need for appropriate morphological handling of words, without which there is no hope for a model to capture enough statistical information about those words.
Moreover, many of these languages have only very small text corpora, substantially magnifying these challenges.
The remainder of this work is structured as follows.

%
%

In \Cref{ch:background} we briefly review the relevant background literature in finite-state morphology, language modelling, and machine translation.
We review finite-state approaches to morphological analysis.
We review the major approaches to language modelling, including $n$-gram language models, feed-forward language models, and recurrent neural language models.
%

In \Cref{ch:resources} we present a set of polysynthetic languages which we will consider throughout this work and detail the resources available for each.
%
%
We have a particular focus on Inuit-Yupik, a highly challenging family of endangered polysynthetic languages that ranges geographically from Greenland through northern Canada and Alaska to far eastern Russia. 
The languages in this family
are extraordinarily challenging from a computational perspective, with pervasive use of derivational morphemes in addition to rich sets of inflectional suffixes and phonological challenges at morpheme boundaries.

In Chapters \ref{ch:mt}--\ref{ch:applications} we examine the current state-of-the-art in language modelling, machine translation, and predictive text completion in the context of four polysynthetic languages: Guaraní, \ess, \esu, and \Iku.
In \Cref{ch:mt} we present experiments and results on machine translation into, out of, and between polysynthetic languages;
we carry out experiments between various Inuit-Yupik languages and English, as well as between Guaraní and Spanish, showing that multilingual approaches incorporating data from higher-resource members of the language family can effectively improve translation into lower-resource languages.
In \Cref{ch:lm}, we present language modelling experiments across a range of languages and vocabularies. 
In \Cref{ch:applications} we present practical applications which we anticipate will benefit from our language model and multilingual approaches, along with preliminary experimental results and discussion of future work.

Finally in \Cref{ch:models} we present a core theoretical contribution of this work: a feature-rich open-vocabulary interpretable language model designed to support a wide range of typologically and morphologically diverse languages. 
This approach uses a novel neural architecture that explicitly model characters and morphemes in addition to words and sentences, making explicit use knowledge representations from finite-state morphological analyzers, in combination with Tensor Product Representations \citep{smolensky1990} to enable successful neural language models capable of handling the full linguistic variety of typologically variant languages.
We present our conclusions in \Cref{ch:conclusion}.

\begin{figure}
    \centering
    \includegraphics[width=\textwidth]{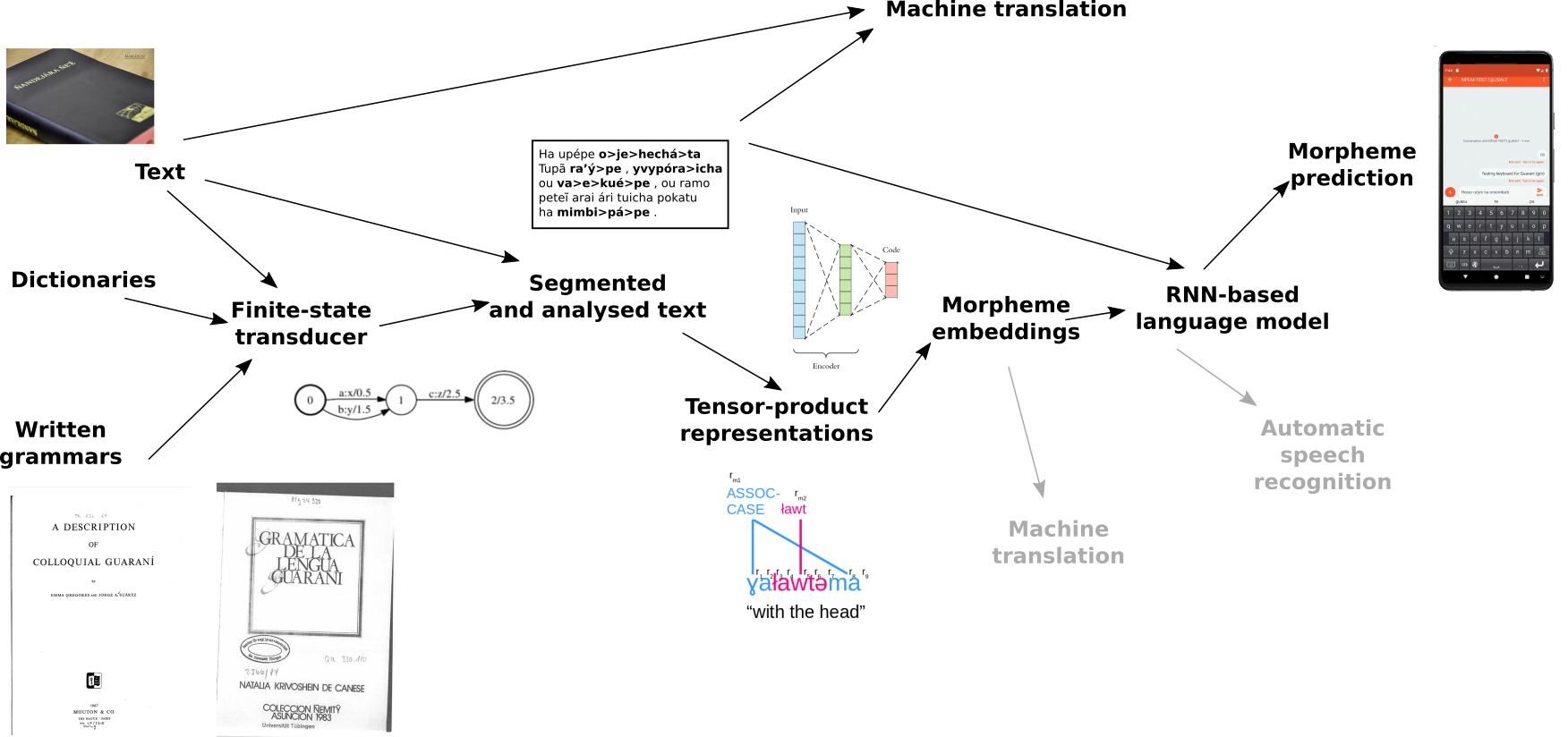}
    \caption{Overview of the tangible artefacts, models, and applications in this report. We start with all of the available
      resources for a given language, including (bi-)texts, grammars, and dictionaries. These are used to create finite-state morphological
      analyzers and MT systems (\S\ref{ch:mt}) directly. The finite-state morphological analyzers are then applied to corpora to create segmented or 
      analyzed corpora (\S\ref{ch:background}).  These are used both to build language models (\S\ref{ch:lm}) and machine translation systems (\S\ref{ch:mt}) based on the segmented morphemes and to create interpretable morpheme-based language models using tensor product representations (\S\ref{ch:models}).
      The final results are 
      predictive keyboards that use morphemes as the unit of prediction (\S\ref{ch:applications}), with potential future work (greyed
      out) including automatic speech recognition and morpheme-based machine translation.}
    \label{fig:overview}
\end{figure}

\chapter{Background\label{ch:background}}



In this chapter we provide a brief overview of the background technologies that underlie this report, namely finite-state approaches to morphological analysis (\S\ref{sec:fsm}), $n$-gram and neural language modelling techniques (\S\ref{sec:LMs}), and neural machine translation (\S\ref{sec:mt-background}).

\section{Finite-state morphology\label{sec:fsm}}

Initial approaches to modelling the morphology of natural languages in the mid-20th century tended to focus on unidirectional algorithmic solutions to particular languages, implemented in general-purpose (rather than domain-specific) programming languages. 
These included 
generators, which generated wordforms from an analysis specification, 
analyzers, which returned possible analyses for a given word, and lemmatizers or stemmers which aimed to return a baseform, stem, or lemma given a wordform.
These approaches had a number of downsides, the first being that the same code could not be used for analysis
and generation, so for each language, separate code had to be written for these two tasks. In addition, descriptions
could not be shared between related languages without much difficulty and there was little formalization.


In the early 1980s this changed with the introduction of finite-state morphology. In this formalization of morphology,
the set of potential strings (wordform-analysis pairs) in a language is represented by a finite-state transducer.
A finite-state transducer is a special class of finite-state automaton where each arc has both an input symbol and
an output symbol.
%
%
%
%
There are two main approaches to modelling morphophonological (or morphographemic) rules using finite-state approaches. The
first consists of applying a sequence of rewrite rules in the form \(\alpha ~ \rightarrow ~ \beta ~ / ~ \gamma ~ \_ ~ \delta\),
where the alphabet symbol \(\alpha\) is rewritten as \(\beta\) between \(\gamma\) and \(\delta\). 
The second approach is referred to as two-level morphology \citep{koskenniemi1983twolevel}. In this approach, phonological
rules are unordered constraints over possible symbol pairs. 
As \cite{karttunen93} notes, the two approaches are formally equivalent and all phonological phenomena that can be 
described with one can be described with the other.

Given a description, a finite-state morphological analyzer can produce both analyses of surface tokens (e.g. sequences
of tags and lemmas such as those found in interlinear glosses) and segmentations of surface tokens. Consider 
the output of the analyzer for the Guaraní sentence \emph{Rehótapa che rendápe.} `Will you come with me' in Example~(\ref{ex:grn-analyzer-output}).
The output includes the lemmas \emph{ho} `come', \emph{che} `my' and \emph{tenda} `place', person and number tags
such as \texttt{<p2>} `second person', \emph{<sg>} `singular', tags indicating word class, \emph{<n>} `noun' 
and \emph{<v>} `verb' among others.

\vspace{5mm}
\begin{tabular}{lll}
\ilg{ex:grn-analyzer-output} & Input & Rehótapa che rendápe. \\
 & Analysis & re<prn><p2><sg>+ho<v><iv>+ta<fti>+pa<qst> \\
 & & che<prn><pos><p1><sg> \\
 & & r<det>+tenda<n>+pe<post> \\
& Segmentation & Rehó>ta>pa che r>endá>pe \\
\end{tabular}
\vspace{3mm}

\noindent
This is especially important for polysynthetic languages, as words can be made up of many morphemes, for example
the word \emph{ñaha'arõ'ỹetéva} `that we did not expect at all' in the sentence \emph{Oiko peteĩ mba'e ñaha'arõ'ỹetéva.} ``Something happened
that we did not expect at all' can be decomposed as in Example~(\ref{ex:longword-grn}) below.

\vspace{5mm}
\noindent
\begin{tabular}{lll} 
\ilg{ex:longword-grn} & Input & ñaha'arõ'ỹetéva \\
& Analysis & ña<prn><p1><pl>+ha'arõ+ỹ<neg>+ete<emph>+va<subs> \\
& Segmentation &  ña>ha'arõ>'ỹ>ete>va \\
\end{tabular}
\vspace{3mm}


The amount of time required to develop a finite-state description can vary widely, but can be anywhere from two weeks,
given a trained developer and a description of a related language --- e.g. Kumyk in \cite{washington:2014} --- to
a year for a developer completely unfamiliar with the tools and language. The speed is also affected by the available
resources such as grammatical descriptions and machine-readable lexicons.

One shortcoming of many finite-state morphological analyzers is an inability to assign probabilities to analyses.
Table~\ref{table:analyses-wound} depicts six example English sentences which each contain the word \textit{wound};
each of these six uses is analyzed with a distinct linguistic analysis.
When analyzing an English sentence that contains the word \textit{wound}, an unweighted English morphological analyzer would posit all of these analyses, and would be unable to suggest which might be the most probable.
%
%
\begin{table}[h]
\centering
\begin{tabular}{llrr}
\toprule
\textbf{Analysis} & \textbf{Example} & \textbf{Frequency} & \textbf{Rel. frequency}\\
\midrule
`wind-{\sc past}' & She wound the watch. & 4 & 0.66\\
`wind-{\sc pp}' & She had wound the watch. & 1 & 0.16\\
`wound-{\sc n.sg}' & The wound healed quickly. & 1 & 0.16 \\
`wound-{\sc inf}' & Therefore I will wound you. & 0 & 0 \\
`wound-{\sc pres}' & They wound and they heal. & 0 & 0 \\
`wound-{\sc imper}' & You wound me sir! & 0  & 0 \\
\bottomrule
\end{tabular}
\caption{List of analyses for the wordform \emph{wound} in English, along with example sentences and frequency
    according to the English treebanks from the Universal Dependencies project \citep{nivre:2016}.}\label{table:analyses-wound}
\end{table}
Some finite-state morphological toolkits support the use of probabilities on arcs in constructed finite-state transducers \citep{mohri:01}.
This means that it is possible to make analyzers and segmenters where the output is ranked, either by probability or by some other metric.
Arc probability weights can be obtained from corpus statistics or from other measures.
This is especially important for polysynthetic languages, where words may potential have many analyses.
We describe the methods we used to weight our analyzers in Section~\ref{sec:preproc-segmentation}.

\section{Language modelling\label{sec:LMs}}



A language model is any model that describes natural language.
By that description, the finite-state models from the previous section could also be considered as a form of language model. 
In this section, however we use a narrower definition of language model as being a model of a probability distribution over a sequence of vocabulary items (characters, words).

Perhaps the simplest approximation to determine the probability of a sentence would be to use a 
unigram model over words. 
In such a model, the probability of a sentence is defined as the product of the probabilities of the individual words, which could be estimated by taking their relative frequency in a given corpus. 
While such a model could reasonably discriminate between the relative probabilities
of sentences such as (a) ``have a great trip'' and (b) ``have a superannuated 
tardigrade'', it would not be able to distinguish the relative probability of 
(c) ``great a have trip'' and (a).
A more accurate, but less tractable approximation would be to ask all speakers of a
given language to rank all of the possible sentences in that language by some metric 
of `goodness'. 
So the idea of language modelling is to find a tractable way to model the distribution
of probability for sequences of linguistic symbols or tokens. 

This simple model can be extended to \(n\)-gram language models \citep{shannon48,shannon51},
whereby instead of modelling single units (characters, words), what is modelled is 
sequences of units. Thus in a bigram word model, the sequences 
modelled would be bigrams, e.g. \{have a, a great, a trip\} and \{great a, a have, have trip\}
from examples (a) and (c) respectively. 
For languages where large amounts of monolingual training data are available, language models of order 5--7 have been widely used in applications such as machine translation and 
automatic speech recognition.

However, as the model is extended to cover longer sequences, the problem of out-of-vocabulary (OOV)
items becomes more severe. This happens when the sequence we are attempting to estimate
the probability of does not appear in our model. This can be illustrated with the example in
(b) above. The sequence ``superannuated tardigrade'' does not return any results with
a search engine query on several major search engines.
It is therefore highly likely that a bigram language model trained using all English text available on the internet would estimate the probability of this sequence
to be zero, and therefore the probability of the entire sentence would also be zero.
%
There are two techniques that have been developed to deal with this problem. Smoothing 
techniques reserve a small amount of the probability mass to distribute to unseen 
\(n\)-grams \citep{good53,jelinekmercer80,katz87,wittenbell91,churchgale91,neyetal94,kneserney95}, %
while backoff techniques allow combinations of lower-order \(n\)-grams to be used to 
estimate the probability of higher-order ones. In example (b) the probabilities of 
`superannuated' and `tardigrade' would be used to estimate the probability of `superannuated
tardigrade'.


One of the issues with \(n\)-gram language models is that parameters are not shared between tokens
and sequences. For example, the token `wonderful' is as far from `great' as is the token
`superannuated'. So if we have the sequence ``have a wonderful trip'', the other shared contexts
that `wonderful' and `great' appear in are not taken into account.
A way of dealing with this problem is to use distributional representations of individual
tokens, as in \citet{2000:Bengio:etal,2003:Bengio:etal}. Here each token is represented by
a vector of real numbers, embedding each token in a shared vector space.
In these kind of language models it is still necessary to specify a fixed \(n\)-gram context,
which means that the amount of context that can be taken into account is limited to a 
fixed-sized window for each token. \cite{2010:Mikolov:etal} describe using
recurrent neural networks to model context to allow whole-sentence context to be taken into 
account. In addition they introduce efficient methods of training the distributional vectors
such that corpora numbering in the billions of words can be used in training.
%
%
%
In both the models proposed by \cite{2000:Bengio:etal} and \cite{2010:Mikolov:etal} each token
is represented by a single vector. As evidenced from the examples above this is not always
tenable, words in natural language are ambiguous (cf. \emph{wound} and \emph{trip} -- `to trip over something'
or `a nice trip').
In ELMo \citep{2018:Peters:etal} and BERT \citep{2019:Devlin:etal}, each word vector is context dependent, both on external, sentence-level
context, and on word-internal context, so even if a given token has not been seen before, 
the model can generalize from forms that have similar surface forms and appear in similar contexts. 
This would seem to be an ideal model for polysynthetic languages, however the downside is that 
these models typically contain very large numbers of parameters which in turn require very large amounts of training data, far more than is available for most endangered languages.




\section{Machine Translation \label{sec:mt-background}}
In recent years, the machine translation community has gravitated toward neural approaches to machine translation.
Midway through the 2010s, these began outperforming phrase-based statistical and other approaches in large-scale evaluations \citep{bojar-EtAl:2016:WMT1}.
This success has driven a rapid sequence of approaches to building neural machine translation models, from sequence-to-sequence models \citep{NIPS2014_5346}, to models with attention \citep{bahdanau2014neural}, to models that primarily rely on attention \citep{vaswani2017transformer}.
In preparation for the workshop, we trained both statistical and neural machine translation models on the available training data.
During the workshop, we focused solely on neural approaches to machine translation, and report those experiments in \cref{ch:mt}.
As our experiments tended to examine variations of the input to the translation models rather than modifications to the networks themselves, we do not provide a thorough overview of the techniques here; for additional detail, please see the cited code and papers.

There does exist prior work on machine translation for polysynthetic languages, though it has generally been limited by small data sizes.
In their recent overview of corpus resources for indigenous languages of the Americas, \citet{mager-etal-2018-challenges} note that most of the parallel corpora they found were quite small (less than 250,000 lines of text).
\citet{homola2012} proposed the use of rule-based systems for polysynthetic languages, but this approach is still labor-intensive, as it requires the application of extensive linguistic knowledge or other tools.
\citet{monsonbuilding} report on Mapudungun and Quechua to Spanish machine translation systems.
\citet{mager-etal-2018-lost} discuss challenges of translating between polysynthetic and fusional languages.
This is not a complete account of all such work.

Of special note for the purposes of this work is existing research on two of the languages we worked on this summer: \Iku{} and \Grn.
For translation between \Grn{} and \Spa{}, we are aware of an online gister (\url{http://iguarani.com/}) and Bible translations evaluated on stemmed output \citep{rudnick-dissertation}, and a system for translators called \emph{Mainumby} by \citet{Gasser:18}.
Previous work on translation between \Iku{} and \Eng{} can be found in 
\cite{micher:2018}, in which results of statistical machine translation for \Eng{} and \Iku{} are reported.  Micher makes use of a morphologically analyzed previous version of the Nunavut Hansard corpus to enhance SMT systems.  Details on developing this corpus can be found in \cite{micher2018techreport}.  The FST-based analyzer \citep{Uqailaut} in combination with the neural analyzer \citep{micher2017improving} are used to morphologically analyze this data set.
\citet{klavans-etal-2018-challenges} discuss some of the challenges of building such translation systems.

%


\chapter{Languages \& Resources\label{ch:resources}}

%
A central issue that arises when conducting research on polysynthetic languages is the lack of resources: many polysynthetic languages are very low resource. 
%
Due to the need for corpora for use in language modelling efforts, 
an effort
was directed towards locating existing corpora for polysynthetic languages and assessing their usability for different experiments.
While we used only a subset of what we collected for experiments, this chapter provides an overview of all linguistic resources we gained access to in the process in order to offer a glimpse into available polysynthetic language resources.

In what follows, we provide short descriptions of the language families and languages involved and the corpora we collected.
We briefly discuss the characteristics of polysynthetic languages based on descriptive statistics and the texts we selected for subsequent experiments.
Details regarding corpus preprocessing are described in the context of experiments discussed in later chapters.

\section{Language selection \& data collection}

We obtained corpora and resources for six languages: Chukchi, \ess, \esu, Inuktitut, Crow, and Guaraní. These languages were chosen from four different families, all of which are low-resource and polysynthetic. There was a focus in particular on the Inuit-Yupik-Unangan family, from which three of the languages were selected.
%
%
The Inuit-Yupik-Unangan languages, historically known as Eskimo-Aleut, are a language family native to the Russian Far East, Alaska, Canada, and Greenland. 
The family is divided into two branches: Inuit-Yupik and Unangan. 
\ess, \esu, and \Iku{} belong to the Inuit-Yupik branch of the family.






In preparation for the workshop, we gathered spoken and written corpora for the selected polysynthetic languages.
%
%
In addition to written and spoken corpora, 
where available, we also gathered dictionaries, reference grammars, and finite-state morphological analyzers. 
Table \ref{lang-overview} provides a summary of the resources we had in each language.
We refer to each language by name or by ISO 639-3 code.

\begin{table}[h]
\centering
\begin{tabular}{llcccc}
\toprule
\textbf{Language} & \textbf{Code} & \textbf{Mono. text} & \textbf{Para.
text} & \textbf{FST} & \textbf{Audio} \\
\midrule
\ckt & \CKT & ✓ &  &  ✓  & ✓ \\
\ess & \ESS & ✓ & ✓ & ✓ & ✓ \\
\esu & \ESU & ✓ & ✓ &  &  \\
\Iku & \IKU & ✓ & ✓ & ✓ &  \\
\cro & \CRO &  &  &  & ✓ \\
\Grn & \GRN & ✓ & ✓ &  ✓  & \\
\bottomrule
\end{tabular}
\caption{\label{lang-overview} Overview of languages and resources: monolingual text, parallel text, finite state transducers, and audio data.}
\end{table}




\subsection{Chukchi}
Chukchi (\CKT) is the most widely spoken language in the Chukotko-Kamchatkan family, with approximately 5000 speakers.
The Chukotko-Kamchatkan languages are native to the Russian Far East, and Chukchi is spoken in the easternmost part, mainly on the Chukotka Peninsula.

We obtained audio data and transcripts for Chukchi from \url{http://chuklang.ru}, a website dedicated to materials and research on Chukchi funded by the Russian Science Foundation.
%
The audio data contains two books of the Bible, the Book of Jonah and the Gospel of Luke, and short stories in the language. The stories represent a valuable resource for the endangered language.
The transcripts are in both Latin and Cyrillic scripts.
There also exists a prototype finite-state morphological analyzer for Chukchi \citep{andriyanets-tyers-2018-prototype}. This analyzer was expanded on during the workshop using the transcripts of the audio data.

\subsection{\ess}


{\ess} (\ESS) is an endangered language in the Inuit-Yupik family spoken on St.~Lawrence Island, Alaska and on the Chukokta Peninsula of the Russian Far East.
We collected a corpus consisting primarily of scanned and digitized books,
%
%
including educational materials  \citep{Apassingok:1993,Apassingok:1994,Apassingok:1995}, oral narratives \citep{Nagai:2001, Apassingok:1985:Vol1,Apassingok:1987:Vol2,Apassingok:1989:Vol3, Slwooko:1977,Slwooko:1979} and a reference grammar \citep{Jacobson:2001}. 
In addition, we made use of 
the Yupik translation of the New Testament%
\footnote{\url{https://live.bible.is/bible/ESSWYI}}
\citep{YupikBible}.
We made use of the \citet{chenschwartz:LREC:2018} finite-state morphological analyzer, which was based on the Yupik grammar of \citet{Jacobson:2001} and incorporated Yupik lexical entries from the \citet{Badten:2008} dictionary.
%
%
%
%
%

\subsection{\esu}

{\esu} (\ESU) is an official language of Alaska that is spoken by about 10,000 speakers in the western and southwestern parts of the state. There are five major dialects of {\esu}, of which General Central Yupʼik (Yugtun) is the most widely spoken.

This workshop made use of a Yupʼik translation\footnote{\url{bibles.org}} of the Bible. 
%
%
As one of our team members speaks the language, we were able to align it with a corresponding English Bible (Good News Translation, Today’s English Version, Second Edition).
The parallel data were used for both machine translation and language modelling experiments. 
Additionally, the Yup'ik Bible and a dictionary \citep{jacobson1984yup} were used to begin development on a Yup'ik finite-state morphological analyzer.

\subsection{Inuktitut \label{sec:iku-data}}

Inuktut (a term that includes the variants \Iku{} and Inuinnaqtun) is one of the official languages of Nunavut, the largest territory of Canada, and is spoken by approximately 39,770 people in Canada \citep{statscanada2017}.
It also has official recognition in several other areas and is part of the Inuit-Yupik-Unangan language family.
Inuktut can be written in syllabics or in roman orthography, and regional variations use different special characters and spelling conventions. 

As Inuktut is an official language of government in Nunavut, there exist some resources that are available in this language at a much larger scale than most other languages in the same family, notably a parallel corpus with \Eng.
Since its formation in 1999, the Legislative Assembly of Nunavut has been publishing its proceedings (known as a Hansard) in both \Iku{} (\IKU) and \Eng.\footnote{It should be noted that Legislative Assembly of Nunavut discourse takes place in several Inuktut varieties, as well as English; a more detailed description of the construction and dialect situation of the Hansard will be available in \cite{v3-hansard}.}
In the subsequent 20 years, the collected Nunavut Hansard has grown to be a substantial bilingual corpus \citep{martin2003aligning,martin2005aligning,nhv2,v3-hansard}, putting Inuktitut in the perhaps unique position of a polysynthetic language with a parallel corpus of more than a million sentence pairs.  
We discuss the different versions of this data, and their preprocessing for machine translation, in \Cref{sec:MT-data}.

We also made use of a Inuktitut translation\footnote{\url{bible.com}} of the Bible for language modelling experiments. 
We decided to exclude the Hansard in the language modelling experiments as including it would make the Inuktitut dataset substantially different from other datasets and thus making it hard to compare it with other languages.
How we preprocessed the data for language modelling is discussed in \Cref{ch:lm}.
\subsection{Crow}
\label{sec:resources:crow}
Crow (Apsáalooke, language code \CRO) is one of the most widely spoken languages of the Siouan family, with approximately 3500 speakers.
The Siouan languages are native primarily to the Great Plains of North America, and Crow specifically is spoken in southern Montana.

Our primary resource for Crow was a series of audio recordings for a dictionary developed by the Language Conservancy, an organization that protects and revitalizes Native American languages. This corpus consists of 11.7 hours of recordings produced by 14 speakers. The data is entirely composed of single words and short phrases from the online Crow Dictionary project \citep{CrowDictionaryOnline}. This data was obtained on special permission from the Language Conservancy and is not publicly available.




\subsection{Guaraní} 

Guaraní (\GRN) is a Tupian language native to South America.
It is an official language of Paraguay and the most widely spoken language in the country with almost 5 million speakers. It is also the only indigenous language of the Americas with a large number of non-indigenous native speakers.

We were able to obtain Guaraní-Spanish parallel Bible translations. The Guaraní Bible was translated and published by the Sociedad Bíblica Paraguaya.
The parallel translations were used for language modelling and machine translation experiments. A morphological analyser developed by \cite{kuznetsova:19}, \texttt{apertium-grn}, was also used.

\begin{table}
\centering
\begin{tabular}{lllrrrrr}
\toprule
\textbf{Language} & \textbf{Code} & \textbf{Corpus} & \textbf{Sentences} & \textbf{Tokens} & \textbf{Types} & \textbf{TTR} & \textbf{MDN}\\
\midrule
\esu & \ESU & Bible & 59575 & 566544 & 138320 & 0.244 & 3.86 \\
\Eng & \ENG & Bible & 62049 & 1057713 & 22201 & 0.021 & 42.90 \\
\ckt & \CKT & Transcripts & 1015 & 5309 & 2387 & 0.450 & 2.22 \\
\Iku & \IKU & Bible & 31103 & 459571 & 126165 & 0.275 &  3.64\\
\Iku & \IKU & Hansard  & 1300148 & 10869995 & 1563883 & 0.144 & 6.95 \\
\Eng & \ENG & Hansard  & 1300148 & 20367595 & 59234 & 0.003 & 343.81 \\
\Grn & \GRN & Bible & 30078 & 629099 & 45766 & 0.073 & 12.71 \\
\Spa & \SPA & Bible & 30078 & 822192 & 31625 & 0.038 & 23.75 \\
\ess & \ESS & Books & 24456 & 214090 & 60414 & 0.282 & 3.32 \\
\ess & \ESS & New Testament & 8002 & 119482 & 32532 & 0.272 & 3.45 \\
\Eng & \ENG & New Testament & 8002 & 273064 & 9071 & 0.033 & 28.37 \\
\bottomrule
\end{tabular}
\caption{\label{lang-overview-stats} Statistics of the written corpora, including type-token ratio (TTR) and mean distance to next novel type (MDN).}
\end{table}

\section{Descriptive statistics of the corpora}



The polysynthetic languages described above differ significantly from languages such as English and Spanish.
One major point of difference is in the ratio of word types to word tokens; 
given the number of word tokens and the number of unique word types, the type-token ratio is calculated as $\textrm{TTR} = \frac{|types|}{|tokens|}$.
Another useful metric, proposed by \citet{hasegawa2017arabic} and used for polysynthetic language by \citet{schwartz-etal-2020-inuit-studies}, calculates the mean distance to the next novel word type (MDN).
\begin{algorithm}[t!]
\KwResult{Mean distance to next novel type}
    distances = list;\
    
    types = list;
    
    current\_distance = 0;\
 \For{word in text}{
 
 \If{word in types}{
 
    current\_distance++;
 
 }
 \Else {
 
    distances.append(current\_distance);
    
    current\_distance = 0;
    
 }
 
 }
 distance = avg(distances)
 \caption{Mean distance to next novel type metric}
 \label{pscode:mean_distance_to_unseen}
\end{algorithm}



Table \ref{lang-overview-stats} displays these text metrics for all textual corpora used. Large differences exist between different languages and between different corpora of the same language with respect to these metrics. The polysynthetic languages examined display higher type-token ratios and lower average distances to the next novel word type in comparison to the non-polysynthetic languages (English and Spanish). This is particularly poignant for parallel corpora. The New Testament in English has a type-token ratio approximately nine times lower than \Ess. This is somewhat expected as the central focus of this work is determining effective strategies for working with highly morphologically complex polysynthetic languages and previous research \citep{kettunen2014} has indicated that morphological complexity is correlated with metrics like TTR.

The datasets utilized cover a large number of different domains as well, including religious texts, parlimentary proceedings, audio transcriptions, and data scraped from internet resources. These domain differences contribute to the differences in corpus properties as well. For example, both the English Bible and the English Nunavut Hansard corpus have lower type token ratios and higher mean distances to the next novel type. However, the formulaic language of parliamentary proceedings causes the English Hansard corpus to have a type-token ratio seven times lower than the English Bible used. These domain differences were controlled for the language modelling experiments described in Chapter \ref{ch:lm} by using the New Testament for several different languages. For the other tasks, comparisons between languages are used sparingly if similar genres of text are not available for both languages.


\section{Preprocessing}
\label{sec:preproc}

We preprocessed the corpora for 1) machine translation and 2) language modelling experiments. 
The general principle and strategies we adapted for preprocessing for both experiments are very similar.
We removed any redundant lines and verse numbers to clean up the corpora.
We made sure to normalize apostrophes so that they remained as part of a word after we tokenized the data using Moses scripts \citep{koehn-etal-2007-moses}.
As truecasing is a common practice in machine translation, we truecased the text for machine translation experiments, but not for language modelling experiments.
Using the cleaned-up datasets, we explored different tokenization strategies. 
FST and BPE segmentation methods were adapted for machine translation experiments, and character, BPE, Morfessor and FST segmentation levels were used for language modelling experiments. 
Details about how we selected and preprocessed the datasets for the two sets of experiments are discussed in \Cref{ch:mt} (Machine Translation) and \Cref{ch:lm} (Language modelling), respectively.

\section{Estimating weights for finite-state morphological analyzers}
\label{sec:preproc-segmentation}

We used three approaches to estimate weights for our finite-state analysers, one supervised, one heuristic and one unsupervised.
The supervised method was the most simple. We had a small corpus of annotated (manually disambiguated) text for
Guaraní, the test corpus from \cite{kuznetsova:19}. We used this and assigned a weight to all wordform:analyses pairs
of \(1\). For the wordform-analysis pairs found in the corpus, a weight was assigned equal to \(1 - P(a|w)\),
where \(P(a|w)\) is the number of times the analysis occurs with the particular wordform over the total number of
times the wordform appears. This is necessarily a number between zero and one and thus for wordforms seen in the corpus,
their analysis received a lower weight than unseen wordform-analysis pairs. Given the size of the corpus, 2020 wordforms,
the majority of the wordforms seen in the corpora were unseen.
For both the Yupik analyser and the Guaraní analyser we added an additional heuristic, for each morpheme boundary,
we increased the weight by \(1\). The motivation behind this heuristic is that we wanted to favor lexicalized forms
and defavor forms with very many derivations when there was a lexicalized alternative.
In addition, we experimented with a novel unsupervised approach to weighting the transducers based on
byte-pair encoding \citep[BPE;][]{sennrich-etal-2016-neural}.
%




 


\newcommand{\seg}{\,$|$\,}

\chapter{Machine Translation\label{ch:mt}}
\section{Introduction}
This chapter discusses neural machine translation (NMT) experiments for translation into, out of, and between polysynthetic languages.
While polysynthetic and, more generally, morphologically complex languages are often considered to pose a greater challenge for machine translation research than languages with relatively simple morphology \citep{OflazerDurgarelkahlout2007,wmt15}, this challenge is often entangled with the challenges of low-resource machine translation.
What really causes this challenge?
Is it the length and complexity of the word forms?
The type-token ratio and data sparsity?
A lack of sufficient training data or a need for more training data than morphologically simple languages?
A matter of many evaluation metrics being ill-suited to morphologically complex languages?
Some combination of all of this?

In this work, we take steps towards answering two relevant questions through experiments on machine translation between \Eng, \Iku, and \Es{} as well as \Grn{} and \Spa.
First, can we untangle the influences of small data and morphological complexity on the challenge of modelling these languages?
Second, can we make use of higher-resource languages in the same language family to improve machine translation of lower-resource languages?
We examine the first through translation of \Iku{} using a new, larger, pre-release version of the Nunavut Hansard,\footnote{While this was a pre-release at the time of this workshop, the data has now been made available publicly; see \cite{v3-hansard}.} as described in \Cref{sec:iku-data,sec:IKU-MT-data,sec:IKU-MT-experiments}.
We examine the second through a series of experiments on low-resource machine translation (described in \Cref{sec:Low-Resource-MT}); our most promising experiments incorporate \Iku{} data into the translation of \Es{} data (\Cref{tab:multilingual}).

We first discuss the data resources for machine translation, providing more detail about data size, preprocessing, and the like (\Cref{sec:MT-data}).
This is followed by descriptions of our machine translation experiments.
\Cref{sec:MT-eval} briefly covers challenges of machine translation evaluation for polysynthetic languages.

The main contributions of our machine translation work during this workshop are as follows.
\begin{itemize}
    \item We achieved state-of-the-art performance on translation between \Iku{} and \Eng{} (since surpassed by \cite{v3-hansard}).
    \item With first access to the beta version 3.0 of the Nunavut Hansard \citep{v3-hansard}, we were able to provide feedback and best practices for preprocessing the dataset and contributed to knowledge about existing character and spelling variations in the dataset. 
    \item We collected empirical evidence on several well-known but unresolved challenges, such as best practices in token segmentation for MT into and out of polysynthetic languages, as well as an examination of how to evaluate MT into polysynthetic languages.
    \item We successfully used multilingual neural machine translation methods to improve translation quality into low-resource languages using data from related languages.
Notably, our ``low-resource'' languages were lower resource than much of the literature, and we produced improvements without the use of large monolingual corpora (which are unavailable for these languages and many other languages of interest).
We observed these improvements across both $n$-gram-oriented and semantic-oriented metrics.
\end{itemize}{}

\section{Parallel Data Resources}\label{sec:MT-data}
\Cref{ch:resources} describes the general data resources used throughout the workshop.
Here we provide a more in-depth look at the resources used for machine translation specifically, including some notes on preprocessing.

\begin{table}[ht]
    \centering
    \begin{tabular}{|c|r|r|r|}
    \hline
    & \multicolumn{1}{c|}{Train} & \multicolumn{1}{c|}{Dev.} & \multicolumn{1}{c|}{Test} \\ \hline
    \IKU-\ENG & 1300148 & 3088  & 2780  \\ \hline
    \ESS-\ENG & 5838 & 1142 & 1750 \\ \hline
    \ESU-\ENG & 30724 & 1279 & 927 \\ \hline
    \GRN-\SPA & 28050 & 1129 & 875 \\ \hline
    \end{tabular}
    \caption{Preprocessed lines of parallel training, development/validation, and test data for machine translation experiments.}
    \label{tab:lrmt-lines}
\end{table}{}

The machine translation resources available to us ranged from moderate to extremely low resource, as shown in \Cref{tab:lrmt-lines}.

\subsection{\Iku--\Eng{} Data}\label{sec:IKU-MT-data}

As described is \cref{sec:iku-data}, there have been several releases of the Nunavut Hansard.
The first, version 1.0, was released to the natural language processing community in \cite{martin2003aligning}, and consisted of 3.4 million \Eng{} tokens and 1.6 million \Iku{} tokens of parallel data.
A subsequent update, version 1.1, corrected some errors in version 1.0 \citep{martin2005aligning}.
Version 2.0 covered proceedings from 1999 through late 2007 (excluding 2003) with about 5.5 million \Eng{} tokens and 2.6 million \Iku{} tokens \citep{nhv2}.

For the purposes of this workshop, we received pre-release access to a beta version of the Nunavut Hansard \Iku--\Eng{} parallel corpus version 3.0, which contains 17.3 million English tokens and 8.1 million Inuktitut tokens, a huge increase over the original data.
We refer to this pre-release version as 3.0 or 3.0 beta.
We use deduplicated development and test sets in our experiments.
The final Nunavut Hansard \Iku--\Eng{} parallel corpus version 3.0 corpus is now available and is described in \cite{v3-hansard}.
Through our early access to this corpus, we provided feedback on the corpus and on preprocessing best practices, which have been incorporated into the data release.

The corpus contains 17.3 million \Eng{} tokens and 8.1 million \Iku{} tokens, spanning 1999 to 2017, a major increase over the version 1.0 and 2.0 releases \citep{martin2003aligning,martin2005aligning,nhv2}.
This is the largest corpus we had access to for this workshop, and is arguably no longer truly ``low-resource'' for machine translation research.
It is, however very domain-specific, and differs in domain from the other parallel corpora we use in our experiments.

As prior machine translation work performed translation on romanized \Iku{} \citep{micher:2018}, we chose to do the same.
We converted \Iku{} data from syllabics as follows: we first applied \texttt{uniconv},\footnote{\texttt{uniconv} is distributed with Yudit: \url{www.yudit.org}} then repaired errors (e.g., incorrectly handled accented French characters in the \Iku{} data) using \texttt{iconv}, then identified and corrected other characters  using a hand-built preprocessing script (including treating word-internal apostrophes as non-breaking characters on the \Iku{} side of the data).\footnote{\citet{v3-hansard} provides slightly updated scripts; we note that neither those scripts nor the ones described here fully conform to spelling and romanization conventions as described in the Nunavut Utilities (\url{www.gov.nu.ca/culture-and-heritage/information/computer-tools}).}

We ran standard preprocessing scripts from Moses \citep{koehn-etal-2007-moses}: punctuation normalization, tokenization, cleaning, and truecasing.
We discuss subword segmentation in \Cref{sec:IKU-MT-experiments}.

\subsection{\Es--\Eng{} Data}\label{sec:ES-MT-data}

We had access to parallel data for two \Es{} languages: \Ess{} (\ESS) and \Esu{} (\ESU).
In both cases, all of the available data was verse-aligned data drawn from the Bible. 
For \Ess, we had access to New Testament data only.
We used Luke for development and validation and used John for testing.
The remainder of the data was used for training.
The data was preprocessed for machine translation experiments as follows: bracketed text was removed from the English data,\footnote{This consisted of rephrasings of entire verses, and was not present in all verses.} apostrophes were normalized in \Ess, and then all data was punctuation-normalized, tokenized, cleaned, and truecased using standard Moses scripts \citep{koehn-etal-2007-moses} with English default settings.

For \Esu, we had access to the full Bible.
For consistency, we still used Luke for development and validation and used John for testing.
The remainder of the data was used for training.
For \Esu, we normalize apostrophes and convert characters with certain diacritics that would otherwise be split by the Moses tokenizer.
Both \Esu{} and its corresponding English translations were punctuation-normalized, tokenized, cleaned, and truecased using standard Moses scripts \citep{koehn-etal-2007-moses} with English default settings.
In the case of \Esu, we performed tokenization without aggressive hyphen-splitting.\footnote{This keeps hyphenated suffixes attached, but has the downside of non-ideal interactions with subword segmentation, occasionally breaking suffixed biblical names into two parts, with the latter attached to the hyphen and \Esu{} suffix.}

\Cref{tab:lrmt-lines} shows the number of lines in the datasets; the \Esu{} training data is more than 5 times larger than the \Ess{} training data.

\subsection{\Grn--\Spa{} Data}\label{sec:GRN-MT-data}
As with the \Es{} datasets, we had verse-aligned parallel Bible data available in \Spa{} and \Grn.
We used Luke for development and validation and used John for testing, with the remaining data used for training.
\Grn{} data was first preprocessed with quotation and apostrophe normalization, along with the removal of paragraph and other symbols that were artifacts of the initial data creation.
\Grn{} and \Spa{} data were then punctuation-normalized, tokenized, cleaned, and truecased using standard Moses scripts \citep{koehn-etal-2007-moses} using \Spa{} defaults.

\section{\Iku{} Machine Translation Experiments}\label{sec:IKU-MT-experiments}
Our \Iku{}-English machine translation efforts were largely concerned with doing initial experiments on the pre-release version of the Nunavut Hansard parallel corpus.  Being substantially larger than previous releases -- to our knowledge, by far the largest aligned parallel corpus of a polysynthetic language to date -- this corpus offered a unique opportunity to try contemporary NMT methods on \Iku{}, and consider whether methods of segmentation like byte-pair encoding \citep[BPE;][]{sennrich-etal-2016-neural} are sufficient to handle a language of this level of complexity.  

In the experiments that follow, our baseline systems -- that is, conventional Transformer \citep{vaswani2017transformer} NMT systems, using BPE and standard hyperparameter settings -- always outperformed the experimental systems (which included special segmentation procedures and multi-source attention).
This suggests that contemporary methods are indeed adequate for processing \Iku{}, although we do not consider the case closed as there are many interesting possibilities for principled segmentation that we have not yet explored.

\subsection{Segmentation experiments}\label{sec:segmentation}

In this set of experiments, we contrast automatic segmentation (by byte-pair encoding) with more morphological segmentations based on human knowledge of \Iku{} morphology, and also consider a simple method of combining them.
We perform our machine translation experiments contrasting these approaches in the \Iku-to-\Eng{} direction.

\subsubsection{Byte-Pair Encoding}

Byte-pair encoding \citep[BPE;][]{sennrich-etal-2016-neural} -- broadly, the segmentation of text at the character-level into larger chunks by compressing the text and using the resulting compression units as word segmentation -- has become a ubiquitous practice in current machine translation.
While the units discovered are not guaranteed to correspond to \emph{morphemes} as such, the resulting systems do end up working at a more morpheme-like level, with units larger than a character but smaller than a word.

\Cref{tab:iku-bpe-example} shows the segmentation of several words according to four BPE vocabulary sizes.
The \Iku{} loanword \textit{siipiisiikkut} (meaning \textit{CBC} or \textit{Canadian Broadcasting Corporation}) is frequent enough in the corpus that at 30000 merges it is represented as a single token.
The word \textit{qimirruvita} (meaning \textit{are we looking at}, as in the context \textit{Are we looking at trying to find out?} or \textit{qimirruvita qaujimanittinnuk}) can be split into the morpheme \textit{qimirru-} (\textit{to scan, to inspect}\footnote{\url{https://uqausiit.ca/node/10333}}) and the verb ending \textit{-vita?} (\textit{are we (3+) ...?}\footnote{\url{https://uqausiit.ca/verb-ending/vita}}); we see that here BPE successfully respects the morpheme boundary at all sizes, segmenting exactly and only along that boundary with a vocabulary of 30000.
For \textit{utaqqivita} (meaning \textit{are we waiting for?}, as in the context \textit{What are we waiting for?} or \textit{kisumik  utaqqivita?}), the story is somewhat different.
Though the word contains the same suffix \textit{-vita?} with the verb root \textit{utaqqi-} (\textit{to wait}\footnote{\url{https://uqausiit.ca/node/12189}}), BPE does not segment the words along the expected morpheme boundaries; the only segmentation that respects them (500) appears to oversegment.
In these examples, we are able to see clear morpheme splits in the surface form, but this is not always the case. In many cases, the underlying forms may undergo phonological changes at the boundaries where two morphemes meet, making it impossible to segment the word such that the resulting units have a uniform representation across all examples of that morpheme.

\begin{table*}[ht]
    \centering
    \begin{tabular}{c|l|l|l}
        \hline
        \textbf{BPE vocab} & \textbf{ siipiisiikkut} & \textbf{qimirruvita} & \textbf{utaqqivita}\\
        \hline
         500 & si\seg i\seg pi\seg i\seg si\seg i\seg kkut  & qi\seg mi\seg r\seg ru\seg vi\seg ta  & uta\seg qq\seg i\seg vi\seg ta \\
         5000 & si\seg i\seg pii\seg si\seg i\seg kkut & qimirru\seg vi\seg ta & utaqq\seg ivi\seg ta \\
         15000 & siipii \seg si\seg ikkut & qimirru\seg vi\seg ta & utaqqivi\seg ta \\
         30000 & siipiisiikkut & qimirru\seg vita & utaqqivi\seg ta \\
        \hline
    \end{tabular}
    \caption{Segmentation of three words according to BPE at four different vocabulary sizes. }
    \label{tab:iku-bpe-example}
\end{table*}

One of our topics of investigation was whether this procedure alone would be sufficient to pre-process \Iku{} for machine translation, whether more sophisticated morphological processing would be necessary, or whether a combination of the two (morphological processing where possible, BPE for the rest) might prevail.

\subsubsection{Morphological Analysis}
The Nunavut Hansard version 1.1 was the starting point for morphological analysis of the larger, later-released corpus (version 3.0).  
As version 1.1 is a subset of the days of debate included in version 3.0, we made use of prior morphological processing of version 1.1 when possible (processing described in \citet{micher2018techreport} and summarized here). 
Every word type of the version 1.1 corpus was processed with the Uqailaut analyzer \citep{Uqailaut}, which provides morpheme segmentation and labeling (including deep representation and morphological category tags).  About 70\%  of the corpus was analyzable by this tool.  
The remaining 30\% was subsequently processed using a neural morphological analyzer, which is trained on a subset of the Uqailaut processed data \citep{micher2017improving}.  Filtering out noise (concatenations of numbers and alphanumerics), we were left with approximately 413K processed word types from  version 1.1 of corpus.

We then extracted the word types from the larger corpus, using the same noise filtering script as with version 1.1 and omitting the word types that had been successfully processed already from version 1.1.  
We ended up with $\sim$1.14M additional types.  
From these another $\sim$9K words were identified as English and removed, yielding $\sim$1.13M types to process.    
However, we note a few differences between these corpora, which affected the processing pipeline.  
First, the romanization scheme performed for version 1.1 of the Hansard is not identical to the romanization we performed on version 3.0 beta. 
In many cases, the resulting romanizations of words match, but in the cases that do not, the morphological analysis needed to be performed anew.  
For example, there are differences in romanization between Hansard versions (e.g. ``lh" vs. ``\&" for the lateral fricative) and between dialects (e.g. ``s" vs. ``h" for a particular phoneme); since Uqailaut presumes ``\&" and ``h", these are substituted before re-processing.
After all of the pre-processing, we followed the same procedure as with version 1.1 of the corpus, first processing what the Uqailaut analyzer would process, and sending the remaining types through the neural morphological analyzer.  
In total, we have 1,548,500 types, processed through one or the other analyzer.

For our work during the workshop, however, we are training and evaluating using only the Uqailaut segmentations (that is to say, without using the neural parser), as the neural parses were not yet finished at the time of these experiments.
We expect that the more complete analyses of the neural parser will have a more positive effect on downstream performance in future experiments.

In the following experiments, the morphologically processed text uses ``deep" forms, in the sense of \cite{micher2017improving}, rather than the surface forms.  Since Uqailaut, and thus the neural generalization of it, only parse surface words into deep forms (and do not generate surface words from deep forms), we present our experiments with different segmentation approaches solely in the \Iku{} to \Eng{} translation direction.

\subsubsection{System configuration}

The model uses a 3-layer encoder, a 3-layer decoder, a model dimension of $512$ and $2048$ hidden units in the feed-forward networks.  
The network was optimized using Adam \citep{kingma2014method}, with an initial learning rate of $\mathrm{1e}{-4}$, decreasing by a factor of 0.7 each time the development set BLEU did not improve for $8000$ updates, and stopping early when BLEU did not improve for $32,000$ updates. 

In addition to the most common automatic MT evaluation metric, BLEU\footnote{BLEU scores were computed using SacreBLEU \citep{post-2018-call}, compared to untokenized but punctuation-normalized references. \texttt{BLEU+case.mixed+numrefs.1+smooth.exp+tok.13a+version.1.4.2}} \citep{papineni-etal-2002-bleu}, we also evaluated our MT experiments using two recently proposed metrics, chrF\footnote{chrF scores were computed against untokenized but punctuation-normalized references using SacreBLEU with \texttt{chrF2+case.mixed+numchars.6+numrefs.1+space.False+version.1.4.2} settings.} \citep{popovic-2015-chrf} and YiSi \citep{Lo2019}, which were shown to correlate better with human judgments on translation quality in English by \citet{wmt19metrics}.

\subsubsection{Results}
\begin{table}[ht]
    \centering
    \begin{tabular}{|c|c|c|c|c|c|}
    \hline
    \IKU{} segmentation & \ENG{} segmentation & BLEU & chrF & YiSi-0 & YiSi-1 \\ \hline
    5000 BPE         & 5000 BPE         & \textbf{27.7} & \textbf{47.1} & \textbf{62.9} & \textbf{70.8} \\ \hline
    Morphological    & 5000 BPE         & 23.3 & 42.5 &  58.2  &  66.1  \\ \hline
    Morph + 5000 BPE & 5000 BPE         & 26.6 & 46.8 &  62.6  &  70.5  \\ \hline
    \end{tabular}
    \caption{Results of \Iku{}-to-English NMT systems as evaluated by BLEU, chrF, YiSi-0 and YiSi-1 .}
    \label{tab:iku-eng-morph}
\end{table}

We compared BPE of various vocabulary sizes to the morphological analysis described above.  In \Cref{tab:iku-eng-morph}, we observe that morphological analysis underperforms BPE across all metrics.

We think this is not due to a problem in the morphological analysis itself (e.g. identifying morphemes incorrectly), but that the process left unanalyzable words intact, whereas BPE manages to segment all words into more manageable pieces.
We therefore also performed a preliminary attempt to combine them, in hopes of combining some of the benefits of true morphological analysis with the statistical advantages of BPE.  First, we took the output of morphological analysis (i.e., the input corpus to the ``Morphological" system in Table \ref{tab:iku-eng-morph}), trained a new BPE model on it, and segmented it according to this model.  Manual inspection of the results of this process suggest that morphemes identified in morphological analysis were typically left intact by BPE -- that is to say, they were identified as units by BPE as well -- and only unanalyzed words were further segmented.

This system also underperformed the BPE-only system, but only by small margins.  
We think that this avenue is still promising, as there are many possible ways to integrate BPE and morphology.  
Many questions remain:

\begin{itemize}
    \item Does one resegment only the unanalyzed words, or all words?  
    \item Does one \emph{train} the BPE model on only unanalyzed words, or all words?  
    \item Do we use surface morphemes or underlying morphemes?  
    \item Do we rejoin underlying forms or keep them segmented?\footnote{\cite{v3-hansard} finds that using underlying forms, but rejoining them before BPE segmentation, gives a performance improvement over deep forms alone in corpus alignment.}
\end{itemize}

Also, as not all the corpus was fully analyzed, more development in neural analysis will probably lead to improvements downstream.

\subsection{Single source and multi-source experiments}
\label{iku-multisource}

One experimental theme we pursued in this workshop was whether multi-source techniques \citep{zoph2016multisource,nishimura2018crosstranslation,doubleCzech2017multisource}, typically used for training MT systems with multiple source languages, could be of value when applied to multiple \emph{representations} of the input text, as a potential way to combine the benefits of two different kinds of analysis.

A recent result in multilingual machine translation \citep{littell2019multisource} suggested that it can be valuable, when training MT on a corpus that has undergone significant processing (in that case, machine translation of the original source into Russian), to attend to \emph{both} the original text and its processed version.  
That is to say, ``attention" in MT makes it possible to avoid having to choose between using the original text or a process that may have been helpful (or may have destroyed useful information); rather, we can allow the model to attend to the results of any stage in the pipeline, and learn for itself which representations to attend to the most.
The above result concerned a pre-processing step that was itself machine translation -- that is to say, this was a ``pivot" system in which L1 is translated to L2, and L2 is translated into L3.  
We were wondering whether the result might also apply for processing steps that were \emph{not} machine translation. 
Would, for example, it be fruitful to attend to two different pre-processings: say, BPE and morphological, syllabics or romanized, etc.?

\subsubsection{System configuration}

The following experiments were performed using the architecture in \citet{littell2019multisource}, a variant of Transformer \citep{vaswani2017transformer} with multi-source attention, implemented in the Sockeye framework \citep{Sockeye:17} for machine translation.

The model uses two 3-layer encoders (one for each source type), a 3-layer decoder, a model dimension of $512$ and $2048$ hidden units in the feed-forward networks.  
The decoder attended to each decoder using ``flat" attention (that is, attending to each and combining the result by simple addition, rather than an additional, hierarchical attention layer).  
The network was optimized using Adam \citep{kingma2014method}, with an initial learning rate of $\mathrm{1e}{-4}$, decreasing by a factor of 0.7 each time the development set BLEU did not improve for $8000$ updates, and stopping early when BLEU did not improve for $32,000$ updates. 

\subsubsection{Results}

As an initial sanity check, we performed two tests of the idea:

\begin{enumerate}
    \item Source 1: BPE vocab size 5000, source 2: BPE vocab size 30000
    \item Source 1: \Iku{} in syllabics, BPE vocab size 5000; source 2: \Iku{} romanized, BPE vocab size 5000.
\end{enumerate}

We did not expect these to show significant gains, but we wanted to make sure the systems did not experience a serious drop in scores.  
Unfortunately, \cref{tab:iku-eng-multi} indeed showed such a performance drop, with the multi-source systems performing very poorly.  

\begin{table}[ht]
    \centering
\begin{tabular}{|c|c|c|}
    \hline
     Source & Target & BLEU \\
    \hline
     \Iku{}, syllabics, BPE 5000 & English, BPE 5000 & 30.3 \\
     \hline
     \Iku{}, romanized, BPE 5000 & English, BPE 5000 & 27.7\\ 
     \hline 
     \Iku{}, syllabics, BPE 5000 + & English, BPE 5000 & 6.3 \\
     \Iku{}, romanized, BPE 5000 & & \\
     \hline 
     \Iku{}, romanized, BPE 5000 + & English, BPE 5000 & 2.5 \\
     \Iku{}, romanized, BPE 30000 & & \\
     \hline
\end{tabular}
\caption{Preliminary multi-source \IKU$\rightarrow$\ENG}\label{tab:iku-eng-multi}
\end{table}

We believe this is because the multi-source source system greatly increases the number of parameters without an associated increase in information in the corpus.  
If we compare this to the positive results in \citet{littell2019multisource}, the difference is that there the introduction of a third language greatly increases the amount of information available to the system: it is not just another view of the same data.
So, rather than continue exploring additional monolingual multi-source setups (e.g., BPE and morphology together), we instead moved on to the multilingual multi-source experiments detailed in \Cref{lrl-multisource}.

\subsection{Challenges in Evaluation of English-to-\Iku{} MT}\label{sec:MT-eval}

For questions of segmentation, we primarily looked at the \Iku{}-to-English direction, since our morphological analyzer was only able to parse, rather than generate.
(That is to say, while we could output segmented, underlying morphemes, we could not, at that time, rejoin them into fluent outputs.)  
For English-to-\Iku{}, we only looked at BPE-based systems, since these can trivially be de-segmented.
In this translation direction, we focused on questions of evaluation because morphologically complex languages pose a challenge in terms of the choice of automatic evaluation metric.

BLEU \citep{papineni-etal-2002-bleu} is a common metric for evaluation of machine translation output given reference translations.
However, because BLEU score is (typically) computed at the word level, an error in a single morpheme is penalized just as harshly as a completely incorrect choice of terminology.
This can be expected to have a particularly detrimental effect when evaluating translation output in morphologically complex languages; even if the system chooses the correct lemma, any errors of morphological inflection will be counted as whole-word errors, decreasing the count of correctly-predicted $n$-grams.
BLEU score could also be computed over byte pair encodings rather than words, but this poses challenges when trying to compare systems built with different vocabularies. 

chrF sidesteps the segmentation issue by first removing whitespace before counting character $n$-grams and computes a precision/recall-balanced score over the character $n$-gram counts. On the other hand, YiSi-0 respects the word boundaries in the MT output but uses the character-level longest common substring accuracy to evaluate the word-level similarities and aggregates the word-level similarity scores into the sentence-level score. These two automatic evaluation metrics based on character-level information would be more suitable for evaluating MT output in morphological complex languages. In fact, \citet{wmt18metrics} showed that chrF correlates the best with human in evaluating Finnish translation output and YiSi-0 correlates the best with human in evaluating Turkish translation output.
However, we think it important to point out that the complexity of Inuktitut morphology is higher than that of Finnish or Turkish and there is no existing work on MT evaluation for polysynthetic languages.
This remains an area for future work.

\subsubsection{System configuration}
The English-to-Inuktitut MT system was built using the same architecture as that of the system mentioned in \cref{sec:segmentation}. We evaluated the system at both word-level and 5k BPE-vocabulary segmentation using BLEU,\footnote{In this table and table \ref{tab:iku-eng-multi}, BLEU scores were computed against untokenized but punctuation-normalized references using SacreBLEU with \texttt{BLEU+case.mixed+numrefs.1+smooth.exp+tok.13a+version.1.4.2} settings.} chrF,\footnote{chrF scores were computed against untokenized but punctuation-normalized references using SacreBLEU with \texttt{chrF2+case.mixed+numchars.6+numrefs.1+space.False+version.1.4.2} settings.} and YiSi-0. Since YiSi-0 is a weighted harmonic mean of precision and recall, we also dissected YiSi-0 into pure precision and recall for further analysis. 

\subsubsection{Results}
First and the foremost, we have to emphasize that MT system scores for different translation directions are not directly comparable. Thus, one should \textit{not} conclude from \Cref{tab:eng-iku-word} that translating \Iku{} into English is an easier task to the opposite direction, or the translation quality of a system in one direction is better than that in the other direction.

\begin{table}[ht]
    \centering
    \begin{tabular}{|c|c|c|c|c|c|c|}
    \hline
    Translation & Evaluation & & & \multicolumn{3}{c|}{YiSi-0} \\   
    direction & unit & BLEU & chrF & weighted-F & precision & recall \\   \hline
    \IKU$\rightarrow$\ENG & word  & 27.7 & 47.1 &  62.9  & 
                            66.2 &  62.1            \\ \hline
    \ENG$\rightarrow$\IKU &  word  & 17.8 & 46.7 &  48.0  & 
                            49.9 &  47.9            \\ \hline \hline
    \IKU$\rightarrow$\ENG & 5000 BPE & 29.5 & 47.4 &  64.1  &
                            67.6 &  63.3            \\ \hline 
    \ENG$\rightarrow$\IKU & 5000 BPE & 13.7 & 46.4 &  56.4  &
                            59.0 &  56.0            \\ \hline
    \end{tabular}
    \caption{Results of English-to-\Iku{} NMT systems as evaluated by BLEU, chrF and YiSi-0 (with pure YiSi-0 precision, i.e. $\alpha$=0.0 and recall, i.e. $\alpha$=1.0 for supplementary analysis).}
    \label{tab:eng-iku-word}
\end{table}

Instead, we would like to point out that there is a notable difference in word-level BLEU scores for the systems in two translation directions because BLEU penalizes systems on failing to correctly inflect a word form equally harshly as choosing an entirely incorrect word; thus MT systems translating into morphological complex languages are expected to achieve lower word-level BLEU scores. A huge difference can also be seen in YiSi-0 scores using word segmentation in evaluation. However, the chrF score difference between the two translation directions is marginal. 

When evaluating translation output at subword unit level, both BLEU and chrF showed a wider score difference when the translation direction was flipped. However, YiSi-0 showed a smaller difference. The contradicting results showed that evaluating translation output in polysynthetic languages itself is a challenging and unsolved research problem.

Without human evaluation on translation output in polysynthetic languages, we could not conclude whether the quality of the English-to-Inuktitut MT system is acceptable or not (or whether it is sufficient for some use cases but not others). 
We hope that future human evaluation of machine translation into polysynthetic languages will provide a basis for the examination of different evaluation approaches, allowing future researchers to select the most appropriate evaluation metrics.

\section{Low-Resource Experiments}\label{sec:Low-Resource-MT}
In keeping with the theme of the workshop, our low-resource machine translation experiments involve neural systems rather than phrase-based ones, despite the fact that they are built from extremely small datasets. 
While we perform our experiments with fairly simple modern neural models and minimal hyperparameter tuning, recent work \citep{sennrich-zhang-2019-revisiting} suggests that careful tuning of hyperparameters can result in NMT systems outperforming statistical machine translation systems even on datasets of around 5000 sentences (comparable to our smaller datasets).

Most of the low-resource machine translation experiments were performed using Sockeye \citep{Sockeye:17}, and the multi-source generalization of Sockeye introduced in \cite{littell2019multisource}.

\subsection{Baselines and Vocabularies}

\begin{table}[ht]
    \centering
    \begin{tabular}{|c|c|c|c|c|c|}
    \hline
    & RNN & Transf. & Transf. & Transf. & Transf. \\
    & BPE & BPE     & Word    & FST     & FST+BPE\\ \hline
    
    \ESS$\rightarrow$\ENG & 4.2 & {\bf 8.4} & 7.3 & & \\ \hline
    \ENG$\rightarrow$\ESS & 3.3 & {\bf 4.4} & 3.5 & & \\ \hline
    \ESU$\rightarrow$\ENG & 10.7 & {\bf 13.9} & 6.5 & & \\ \hline
    \ENG$\rightarrow$\ESU & {\bf 5.4} & 5.3 & 3.3 & & \\ \hline
    \GRN$\rightarrow$\SPA & & {\bf 10.5} & 7.4 & 7.1 & 9.6 \\ \hline
    \SPA$\rightarrow$\GRN & & {\bf 8.6} & 7.1 & 8.3 & 8.1 \\ \hline
    \end{tabular}
    \caption{BLEU scores of baseline and vocabulary experiments for \Es--\Eng{} and \Grn--\Spa{} machine translation experiments. All BPE vocabularies in this table are of size 5000, learned separately.}
    \label{tab:mt-baselines}
\end{table}

We first compare RNN and Transformer translation models using BPE vocabularies of 5000.
The size of 5000 was selected for consistency with other experiments and because it was among the highest performing vocabulary size on initial RNN experiments for several language pairs (not reported here).
The RNN models were trained using OpenNMT \citep{klein-etal-2017-opennmt} with default settings, and the Transformer models were trained using Sockeye \citep{Sockeye:17} with a 3 layer encoder, 3 layer decoder, batch size 2048, optimized toward perplexity, and the remaining parameters set to defaults.
As \Cref{tab:mt-baselines} shows, the Transformer system outperformed the RNN system in all but one case (which was within 0.1 BLEU); we use the Transformer system for all remaining experiments.

We compare using a BPE vocabulary of 5000 symbols to using a whole word vocabulary.
In all cases, the BPE vocabulary outperforms the whole word vocabulary (by between 0.9 and 7.4 BLEU points).
Using whole words, \Eng--\Ess{} experiments were run with vocabulary sizes of 4787 and 26888 (respectively, including special characters), while \Eng--\Esu{} whole word vocabularies consisted of 13501 and 106736 types respectively.
Given the small data sizes and large \Es{} vocabulary sizes, it is unsurprising that BPE outperforms whole words; there may simply not be enough examples of many types in the long tail for the system to accurately translate them, and the word system includes a large number of out of vocabulary items in the test set.

Following the results of the \Es{} experiments, we omit the RNN experiments for \Grn--\Spa{} and instead start with a baseline of a Transformer model (3 layer encoder, 3 layer decoder, batch size 2048, optimized toward perplexity, remaining parameters set to defaults), using separately learned BPE encodings for \Spa{} and \Grn{} with vocabularies of 5000 types each.
There does exist other work on machine translation for \Grn--\Spa{}, notably an online gister\footnote{\url{http://iguarani.com/}} and work in \cite{rudnick-dissertation}.
Though \cite{rudnick-dissertation} also performs experiments on Bible translation, we do not compare directly, as those results are measured on stemmed output.

For \Grn--\Spa{}, we also experiment with full-word vocabularies, FST-segmented vocabulary (\Grn{} side only; \Spa{} side BPE 5000), and an FST-segmented vocabulary with backoff to BPE (all \Grn{} words left unsegmented by the FST were segmented by a BPE model learned for a BPE 5000 vocabulary on \Grn{}; \Spa{} side BPE 5000).
As shown in \Cref{tab:mt-baselines}, the baseline BPE model outperforms all other experiments.\footnote{BLEU scores were computed against untokenized but punctuation-normalized references using SacreBLEU with \texttt{BLEU+case.lc+numrefs.1+smooth.exp+tok.13.a+version.1.3.7} settings.}

\subsection{\Es{} Language Experiments}
Our \Es{} language experiments begin with baseline RNN and Transformer models.
Finding that the Transformer strongly outperforms the RNN (\Cref{tab:mt-baselines}), we perform the remainder of the experiments with the Transformer architecture only.

In addition to the baseline, we perform two experiments: multi-source experiments on a multi-parallel subset of the data and multilingual NMT system experiments.
BPE vocabularies of size 5000 were learned separately on each language's training data using {\tt subword-nmt} \citep{sennrich-etal-2016-neural}.
Our most promising low-resource experiments, described in \Cref{sec:multilingualnmt} involve the use of higher resource languages from the same language family to build multilingual neural machine translation systems which can then be finetuned for specific low-resource languages.

\subsubsection{Multisource}\label{lrl-multisource}

In order to experiment with multisource machine translation, we build a multiparallel verse-aligned corpus from the intersection of all available \Es{} Bible data.
The resulting New Testament corpus has 5449 lines for training, 1091 lines for development and validation, and 874 lines for testing.
It contains data in \Ess{} and \Esu, as well as data from two \Eng{} Bibles.
We call the \Eng{} Bibles \ENS{} (for the \Eng{} Bible originally aligned to \Ess) and \ENU{} (for the \Eng{} Bible originally aligned to \Esu).
We preprocessed them identically to the baseline experiments, with one change: we removed verse numbers from \Esu{} and its corresponding \Eng{} (\ENU) as those were not present in the \Ess{} corpus.

We compared single-source (Sing.) and multi-source (Mult.) approaches, as described in \S\ref{iku-multisource}, as well as separately learned and jointly learned 5000 symbol BPE representations (the joint BPE representations were learned across all 4 sides of the multiparallel corpus).
For the multi-source experiments, we tried translating into \Esu{} using its corresponding \Eng{} and \Ess{}, as well as translating into \Ess{} using its corresponding \Eng{} and \Esu.
Without any major parameter search, we found that the joint BPE single-source systems performed the best.

\begin{table}[ht]
    \centering
    \begin{tabular}{|c|c|c|c|c|c|}
    \hline
    & Sing. & Mult. & Sing. & Mult. & Mult. \\
    & & & (joint) & (joint) & (joint+tied) \\ \hline
    \ESS--\ESU & 3.8 & & 4.7 & & \\ \hline
    \ENS--\ESU & 4.8 & & 4.9 & & \\ \hline
    \ENU--\ESU & 3.6 & 3.2 & 3.9 & 2.8 & 3.1 \\ \hline \hline
    \ENS--\ESS & 4.0 & 3.1 & 4.4 & 3.4 & 3.4 \\ \hline
    \ENU--\ESS & 4.7 & & 5.4 & & \\ \hline
    \ESU--\ESS & 4.2 & & 4.8 & & \\ \hline
    \end{tabular}
    \caption{BLEU score results for experiments on joint and separate BPE learning, along with multisource experiments. Tested on the multiparallel subset of \Es{} corpora.}
    \label{tab:multiparallel}
\end{table}

As these BLEU scores are extremely low, it is quite difficult to draw any conclusions from this set of experiments; the following notes should be understood in that context.
We do observe that for single-source, using a jointly trained BPE vocabulary performs better than separately trained BPE vocabularies.
This may be due in part to improved translation of copied terms (e.g., names).
We do not observe the same consistency in multisource.
Perhaps unintuitively, in single-source experiments, we find that swapping the \Eng{} Bibles (translating \ENU{} into \ESS{} and \ENS{} into \ESU{}) performs better than the ``correct'' pairs.
This highlights several challenges of performing machine translation using Bible corpora: we do not have a guarantee in our case that the ``source'' \Eng{} Bible is the version from which the \Es{} Bibles were translated, Bible translations may rely on metaphor or other non-literal phrases, and verse alignment provides additional challenges due to mismatches between sentence and verse boundaries.
In some cases, we observe that a sentence spans more than one verse, with a name appearing in the first verse in \Eng{} and in the second verse in \Es{} or vice versa, an impossible challenge for machine translation without extrasentential context to overcome; this is a known challenge in parallel Bible corpora \citep{mayer-cysouw-2014-creating}.
We also did not perform hyperparameter optimization due to time constraints; more extensively tuned models may show different results.

\subsubsection{Multilingual}\label{sec:multilingualnmt}

Multilingual neural machine translation has been proposed as a means of improving neural machine translation of low-resource languages, using a variety of distinct approaches.
These approaches depend are split into approaches to translate into or out of low-resource languages.
\cite{neubig-hu-2018-rapid} explore the multilingual translation task translating from multiple low-resource languages into a single high-resource language.
\cite{gu-etal-2018-universal} also work in the same translation direction, and incorporate large amounts of monolingual data and many closely-related source languages.

Our interest is on translation \textit{into} low-resource languages.
In that direction, \cite{multilingual-wordlevel} perform multilingual neural machine translation by tagging each subword with a language-specific tag, and then building a system based on available training data.
\cite{google-nmt} use a single special token at the beginning of input sentences to indicate the desired target language to translate into.
\cite{rikters-etal-2018-training} follow this approach to do multilingual translation into and out of morphologically rich languages, though their low-resource setting consists of more than 3 million sentence pairs.

\Ess, \Esu, and \Iku{} belong to the same language family.
Despite this, they have very limited vocabulary overlap in our parallel data (less than 1\% type overlap between \Iku{} and \Es{}, and less than a 3\% type overlap between \Ess{} and \Esu).
This is certainly due in part to the different domains we had available: legislative text (\Iku) and Bible (\Es).
As described in \Cref{sec:ES-MT-data} and \Cref{sec:IKU-MT-data}, our data spans a wide range in terms of size, from approximately 5000 lines of text to approximately 1.3 million lines.
We approximately follow the \cite{google-nmt} approach in our approach to translating from English into \Iku{} and \Es{} languages.

\begin{table}[ht]
    \centering
    \begin{tabular}{|c||c||c|c|c|}
    \hline
    & Baseline & Multilingual & \ESS-Ad. Multi. & \ESU-Ad. Multi.  \\ \hline
    \ENG--\ESS & 4.4 & 5.8 & {\bf 6.5} & 1.3 \\ \hline
    \ENG--\ESU & 5.3 & 5.7 & 1.9 & {\bf 6.0} \\ \hline
    \end{tabular}
    \caption{BLEU scores for experiments on multilingual neural machine translation. The baseline is the original Transformer baseline for each language pair. Multilingual is the single multilingual system (trained on \Iku{} and \Es{} data), and the remaining two columns show that system fine-tuned on a particular variety of \Es.}
    \label{tab:multilingual}
\end{table}{}

\begin{table}[ht]
    \centering
    \begin{tabular}{|c||c||c|c|c|}
    \hline
    & Baseline & Multilingual & \ESS-Ad. Multi. & \ESU-Ad. Multi.  \\ \hline
    \ENG--\ESS & 26.9 & 28.0 & {\bf 30.1} & 10.5 \\ \hline
    \ENG--\ESU & 31.0 & 32.5 & 16.7 & {\bf 33.2} \\ \hline
    \end{tabular}
    \caption{YiSi-1 scores (higher is better) computed using \ESS{} or \ESU{} BPE 5000 embeddings built by \texttt{word2vec} \citep{w2v} for experiments on multilingual neural machine translation. The baseline is the original Transformer baseline for each language pair. Multilingual is the single multilingual system (trained on \Iku{} and \Es{} data), and the remaining two columns show that system fine-tuned on a particular variety of \Es.}
    \label{tab:multilingual-YiSi}
\end{table}{}

We train joint BPE (vocabulary 5000) on \Iku, \Ess, and \Esu, downsampling the \Iku{} and upsampling \Ess{} to match the size of \Esu.
We prepend a language tag (e.g. ``$<$\ESS$>$'') to each source and target sentence in the three sub-corpora.
Next we train a Transformer model (our ``multilingual baseline'') on the concatenation of all available training data (with no sampling, 3 layer encoder, 3 layer decoder, 512 embedding size, early stopping on perplexity of the concatenated development data).
For \Ess{} and \Esu, we then fine-tune the multilingual baseline on all language-specific training data (with early stopping based on perplexity on the language-specific development data).
The BLEU score results are shown in \Cref{tab:multilingual}.
\Cref{tab:multilingual-YiSi} reports YiSi results, which follow the same trend as the BLEU score results.
As expected, fine-tuning on language specific data boosts performance on that particular language (while the output on the other language appears to exhibit catastrophic forgetting \citep{Kirkpatrick3521}), giving us our best performance.
However, with BLEU scores in the single digits, it is clear that there is still a long way to go before the MT output may be genuinely useful (e.g. in post-editing or interactive translation) for these low-resource languages.


\chapter{Language Modelling\label{ch:lm}}





In this chapter, we report on language modelling experiments, comparing different tokenization strategies for polysynthetic languages. 
We trained a state-of-the-art RNN language model using the character, BPE, Morfessor and FST as the unit for segmenting text data. 
In order to facilitate comparisons across the tokenization strategies, we carefully selected datasets for two experimental settings: 
1) A setting where all the data available for a language is used and  
2) a setting where only the New Testament in a language is used. 
The former setting provides us an opportunity to utilize all the data we have in a language while the latter allows us to draw a more precise comparison across languages. 
We use the average perplexity per character or the character-level perplexity as a metric to compare different models.
The results show that the linguistically-oriented, FST segmentation strategy performed the best in modelling polysynthetic languages when it was available.
In addition, difficulty of modelling different languages is compared using the average perplexity per word or the word-level perplexity.
The potential of FST in aiding language modelling of polysynthetic languages and implications on comparing models for different languages are discussed.
%
%

\section{Data Preparation}

After much consideration, we selected four low-resource, polysynthetic languages for our language modelling experiments (hereafter referred to by ISO 639-3 code): {\ess} ({\ESS}), {\esu} ({\ESU}), \Iku{} ({\IKU}) and {\Grn}  ({\GRN}).
These languages were chosen because we had the most available text data in them. We had at least the Bible, the Gospel books in New Testament in particular, in these languages, and that allowed us to have a commonality among the datasets to facilitate comparison across the languages.
In addition to the polysynthetic languages, we included two well-researched, non-polysynthetic languages: {\Eng} ({\ENG}) and {\Spa} ({\SPA}).
The {\ENG} and {\SPA} data were included to provide comparison between polysynthetic languages and non-polysynthetic languages as {\ESU} and {\ENG}
and {\GRN} and {\SPA} were parallel translations.

\begin{table}
\centering
\begin{tabular}{lll}
\toprule
Split & Setting 1: All data & Setting 2: New Testament\\
\midrule
Train & Rest of the data available & Rest of New Testament\\
& (e.g. Old Testament, transcripts, stories) &\\
Dev & Luke  & Luke \\
Test & John & John \\
\bottomrule
\end{tabular}
\caption{\label{data-split} Train-dev-test split}
\end{table}

We designed two experimental settings to fully utilize available data while ensuring comparability across different languages. 
As for the 1) \textit{all data} setting, we included any available monolingual data in a given language, including but not limited to the New Testament.
The second setting, the 2) \textit{New Testament only} setting, focused only on the New Testament data in order to further ensure comparability given the near-parallel data across different languages.
Regardless of the settings, Luke was used as the development set and John as the test set to further facilitate fair comparison as we had the Gospel books in all languages.
This ensured that different languages shared a development set and a test set and a part of the train set (the rest of the New Testament) in common even though the exact train set available in each language may differ from one another. 
The train set in the 1) \textit{all data} setting included the New Testament, but may also include the Old Testament, transcripts and oral narratives if available.
This setting, therefore, fully utilizes the data we had in each language.  
In the 2) \textit{New Testament} setting, the development and test sets stayed the same, but the train set included the rest of the New Testament only.
It should be noted that we did not align the Bibles at the sentence level, and there was some variability among different Bible translations as discussed in \cref{ch:mt}.
However, {\ESU} and {\ENG} and {\GRN} and {\SPA} Bible translations were assumed to be parallel, and we assume that the other Bible translations provide comparable texts with similar intensions overall.
While the 2) \textit{New Testament only} setting may provide a more precise comparison, the 1) \textit{all data} setting may be more representative of the reality given the limited size of the data for the former setting.
Table \ref{data-split} summarizes the two experimental settings and the dataset split.

Given the data split, we preprocessed the datasets systematically to further ensure comparability among subsequent language models.
We removed redundant, bracketed texts when applicable, and normalized apostrophes as they were meaningful in some languages and should not be tokenized separately from their surrounding words.
Then, we normalized the punctuation and tokenized the texts using Moses scripts \citep{koehn-etal-2007-moses} with default settings.
The overall preprocessing for language modelling experiments resembles that for machine translation experiments discussed in \Cref{ch:mt} except that we did not truecase the data for langauge modelling experiments. 

Tables \ref{data-split-descriptives-1} and \ref{data-split-descriptives-2} summarize descriptive statistics of the preprocessed data under each setting.
Overall, it seems that the characteristics of a language as captured by the statistics are quite similar under the two settings.
This may not be surprising given that the two settings concern very similar domains.
While it remains to be seen if these descriptive statistics would be similar under a different setting for the languages, we observed the followings for the languages given our datasets:
As discussed in \Cref{ch:resources}, the languages seem different in the TTR and mean distance to the next unseen word.
{\ESS}, {\ESU} and {\IKU} consistently show a higher TTR and a lower mean distance to the next unseen word than {\GRN}. 
While {\GRN} is considered as a polysynthetic language, it seems that {\GRN} might be slightly different from the other polysynthetic languages spoken in Alaska ({\ESS}, {\ESU}, {\IKU}). 
Still, {\GRN} is distinctive from {\SPA} and {\ENG} in that it still had a higher TTR and lower mean distance to the next unseen word.
While the {\SPA} data seems more complex under the New Testament setting, {\ENG} and {\SPA} are consistently simpler than polysynthetic languages in terms of TTR and mean distance to the next unseen word.

It is noted that, across languages, the datasets are similar in terms of sentence counts within each experimental setting. 
While {\ESU}-{\ENG} and {\GRN}-{\SPA} differed slightly in terms of the exact sentence count, they are aligned at the verse level.
The rest of the data are not aligned at the verse level, but they seem to contain similar number of sentences under the respective data conditions.
Note that we did not include the Hansard data for {\IKU}. 
We exclude the data because including it would increase the amount of available data and genre variability for the particular language too much to allow comparison across languages.

Given the similar number of sentences present in each dataset, it is noteworthy that the word count and type count are distinct across the languages.
Again, {\ESS}, {\ESU} and {\IKU} seem similar to each other in that they have a smaller number of words and a large number of types than others. 
This reflects their typological characteristic, that they tend to have longer words with more morphemes, which may lead to more unique tokens.
{\GRN} still seems distinct from the other polysynthetic languages in that the datasets in the language tend to have more words and less unique words. 
In fact, {\GRN} seems to have similarity with {\SPA} in terms of the descriptive statistics even though {\GRN} still has a lower mean distance to the next unseen word than {\SPA}.
{\ENG} seems to be clearly more analytic than the other languages as it has more word counts and less type counts.


\begin{table}
\centering
\begin{tabular}{lrrrrr}
\toprule
Language & Sentences & Words & Types &	Type/Token & Mean distance to unseen\\
\midrule
{\ESS} & 20,899 & 206,691 & 58,637 & 0.28 & 3.28 \\
{\ESU} & 33,102 & 474,499 & 106,381 & 0.22 & 4.15 \\
{\IKU} & 31,103 & 466,705 & 126,162 & 0.27 & 3.70 \\
{\GRN} & 30,078 & 622,999 & 38,944 & 0.06 & 14.63 \\
{\ENG} & 21,835 & 395,368 & 11,258 & 0.03 & 31.72 \\
{\SPA} & 30,078 & 840,937 & 24,829 & 0.03 & 30.39 \\
\bottomrule
\end{tabular}
\caption{\label{data-split-descriptives-1} Descriptive statistics for setting 1 (all available data)} 
\end{table}

\begin{table}
\centering
\begin{tabular}{lrrrrr}
\toprule
Language & Sentences & Words & Types &	Type/Token & Mean distance to unseen\\
\midrule
{\ESS} & 7,860 & 121,549 & 31,928 & 0.26 & 3.57 \\
{\ESU} & 8,464 & 108,757 & 30,980 & 0.28 & 3.32 \\
{\IKU} & 7,858 & 110,977 & 36,573 & 0.33 & 3.03 \\
{\GRN} & 7,896 & 171,350 & 12,779 & 0.07 & 12.20 \\
{\ENG} & 7,870 & 210,395 & 5,067 & 0.02 & 38.82 \\
{\SPA} & 7,896 & 206,707 & 11,371 & 0.06 & 16.01 \\
\bottomrule
\end{tabular}
\caption{\label{data-split-descriptives-2} Descriptive statistics for Setting 2 (New Testament only)}
\end{table}

\section{Tokenization strategies}
We considered five different tokenization strategies in modelling the languages: word, character, BPE, morfessor and FST segmentation methods. In what follows, we briefly explain each tokenization strategy and why they might be helpful in segmenting polysynthetic languages.

\subsection{Word}

A common tokenization strategy is to tokenize text by whitespace or by words. 
While it may be simple and seem intuitive, this tokenization strategy faces data sparsity and out-of-vocabulary (OOV) issues. 
For example, if we tokenize by words, \textit{dog} and \textit{dogs} will count as two separate tokens even though there is much shared information between the two. 
If the train set includes only the singular form and the test set contains only the plural form, the plural form in the test set will be considered as OOV.

\vspace{5mm}
\noindent
\begin{tabular}{llllll}\label{ex:yupik:woman}
\ilg{ilg:aghnaaguq} & \multicolumn{4}{l}{aghnaaguq} \\
& aghnagh & -$\sim$:(ng)u & -$\sim_\text{f}$(g/t)u- & -q \\
& woman & -to.be & -\textsc{Intr.Ind} & -\textsc{3sg} \\ \\
& \multicolumn{4}{l}{`she is a woman'}
\cite[p.25-26]{Jacobson:2001}
\end{tabular}
\vspace{3mm}

This tokenization method is particularly problematic for polysynthetic languages given their rich morphology. 
A word in polysynthetic languages may contain several morphemes to express a sentence-like intension. 
For example, a word in {\ESS}, \textit{aghnaaguq}, consists of four morphemes and is translated as \textit{`She is a woman'} as shown in \Cref{ilg:aghnaaguq}.
Importantly, this results in a high rate of hapax legomena (words appearing only once), which results in much higher OOV rates than observed in most non-polysynthetic languages.
In modelling polysynthetic languages, the word-level tokenization is too unrealistic to be useful in predicting the next word, and its performance may be over-estimated or under-estimated depending on how we reward or penalize OOVs.
For example, if we do not penalize a model for predicting an OOV symbol for the next word, it may predict an OOV symbol repeatedly for a polysynthetic language to falsely record a good performance.
If we do want to penalize OOV, we will have to come up with a metric that does that fairly given our data.
Given that the model we adapted did not penalize OOV, we opted to use language models that would not over-generate OOVs.

\subsection{Character}
One possible solution to such issues of word-level tokenization is to tokenize text by the character.
The character-level tokenization rarely has OOV issues because a text typically consists of a finite set of characters regardless of its morphological complexity.
However, this tokenization method, again, cannot fully utilize the linguistic information present in a text as it reduces all words into a sequence of a finite set of characters.
The relationship between \textit{dog} and \textit{dogs} may be easily captured by a character-level model, but words with more complex morphology like \Cref{ilg:aghnaaguq} may be hard to model using the character as the tokenization unit.

While we report our results for character-level models as the baseline to compare other results to, we note that character-level models may not be meaningful for downstream applications for polysynthetic languages such as keyboard prediction:
Predicting a character at a time when a word consists of several morphemes and a long sequence of characters may be too slow or too low-quality.

\subsection{BPE}
If word-level tokenization is too coarse-grained and character-level tokenization is too fine-grained, it may mean that we need to utilize subword units to segment our data.
As discussed in \cref{sec:segmentation}, byte pair encoding \citep[BPE;][]{sennrich-etal-2016-neural} is a unsupervised segmentation method that uses subword units.
Originally a data compression algorithm \citep{Gage:1994:NAD:177910.177914}, BPE has become one of the standard techniques in neural machine translation since \citet{sennrich-etal-2016-neural}. 
Tokens segmented by BPE can represent texts with the minimum entropy by the fixed vocabulary size, which should be chosen as the hyperparameter. 
BPE segmentation may look like morpheme segmentation for some words, but it is data-driven rather than based on linguistic information. 
For example, with enough support from a given data, BPE may segment `lower' as `low@@ er' (@@ represents a within-word morpheme boundary), which may seem linguistically motivated, but it is also possible to get different segmentations such as `l@@ ow@@ er' with different hyperparameters and different data conditions. 
Refer to \Cref{tab:iku-bpe-example} for examples of BPE segmentations for machine translation of {\IKU}, some of which respect morphological boundaries and some of which do not.

We trained a BPE model on the training data and applied the model to all data using \texttt{subword-nmt}\footnote{\url{https://github.com/rsennrich/subword-nmt}}.
We experimented with different vocabulary sizes for BPE segmentation, and report results on two vocabulary sizes: 500 and 5,000. 
While BPE provides an off-the-shelf method to segment words into subword units, it remains unclear whether the unsupervised method would prove useful in modelling polysynthetic languages.

\subsection{Morfessor}

We adopted another unsupervised segmentation method called Morfessor to compare with BPE. 
Morfessor is a tool for unsupervised (and semi-supervised) morphological segmentation and has been utilized in speech recognition, MT, and speech retrieval. 
While there is no literature on its efficiency in neural language modelling tasks for polysynthetic languages, it is said to be useful in modelling languages with rich morphology such as Finnish, Estonian, German and Turkish \citep{MorfessorToolkitStatistical2014}.
Morfessor uses Maximum a Posteriori (MAP) estimation to approximate morpheme segmentation assuming that a word consists of one or more ``morph'', yet its results may not be the same as linguistically motivated morpheme segmentation. 
We used Morfessor 2.0 with the default settings for Morfessor segmentation.

\subsection{FST segmentation}

The last segmentation strategy we considered was segmentation based on FSTs. FST segmentation provides knowledge-based, rule-based segmentation based on linguistic knowledge and analysis. 
Several FST-based morphological analyzers or morphological segmenters have been developed for polysynthetic languages, and we were able to experiment with two of them for our experiments: {\ESS} \citep{chenschwartz:LREC:2018} and {\GRN} \citep{kuznetsova:19}. 
The FST-based morphological analyzers produce zero or more morphological analyses for any given word. 
When there are more than one analysis available for a word, we used heuristics (e.g. choose the shortest
 analysis) to select one analysis to segment the given word. 
When there was no analysis available, we used character (character backoff) or BPE (BPE backoff) segmentation for the word.
The BPE backoff was performed using the existing BPE segmentations with the vocabulary size of 500 and 5,000.
While we were able to obtain this segmentation results only for two polysynthetic languages, this provides a point of comparison between supervised, linguistically motivated segmentation and unsupervised, data-driven segmentation.

\section{RNN-LSTM}
We used a state-of-the-art language model \citep{merityRegOpt, merityAnalysis} for our language modelling experiments. 
The RNN model with LSTM has shown to be competitive in modelling English benchmark datasets such as PTB and WikiText-2.
We adapted the hyperparameters for WikiText-2 (WT2) with LSTM for Morfessor and FST models and the hyperparameters for character level enwik8 for character and BPE models. 
\Cref{lm-hyperparameters} summarizes the hyperparameters.

We acknowledge that 
none of these models (nor any other models to our knowledge) have been specifically designed to model polysynthetic languages or reported to be used to model polysynthetic languages.
With a lack of a language model designed to model polysynthetic languages, we chose a state-of-the-art model that has proven competitive in modelling English instead.

\begin{table}[t!]
\centering
\begin{tabular}{lrr}
\toprule
& Character \& BPE & Morfessor \& FST \\
\midrule
RNN Cell & LSTM & LSTM \\
Layers & 3 & 3\\
RNN hidden size & 1840 & 1150\\
Dropout (e/h/i/o) & 0/0.1/0.1/0.4 & 0.1/0.2/0.65/0.4\\
Weight drop & 0.2 & 0.5\\
Weight decay & 1.2e-6 & 1.2e-6\\
BPTT length & 200 & 70\\
Batch size & 128 & 80 \\
Input embedding size & 400 & 400 \\
Learning rate & 1e-3 & 30\\
Epochs & 50 & 200 \\
Random seed & 1111 & 1882 \\
Optimizer & Adam & SGD \\
LR reduction (lr/10) & [25, 35] & NA\\
\bottomrule
\end{tabular}
\caption{\label{lm-hyperparameters} Hyper-parameters for word- and character-level language modelling experiments}
\end{table}

\section{Character-level perplexity}
Perplexity is a measure of language modelling difficulty and calculated by taking the exponent of the average negative log-likelihood per token.
Because perplexity as it is depends on the tokenization strategy, we calculate the character-level perplexity for each model to allow comparison among them. 
We define the character-level perplexity as the exponent of the average negative log-likelihood per character and calculate it by adding up the token-level loss for a given tokenization, multiplying the total loss by the number of tokens in the test set and dividing the value by the number of characters in the test set.
We count whitespace and the end of a sentence symbol as separate tokens.
This ensures a fair comparison among different tokenization strategies. 
The choice of character as the common denominator is arbitrary, and it can be other tokenization methods such as the word. 
Refer to \citet{compare-perplexity} for detailed explanations.

\section{Results \& Discussion}

\begin{table}[t!]
\centering
\begin{tabular}{lrrrr}
\toprule
Language & Morfessor & BPE (V=500) & BPE (V=5k) & Character\\
\midrule
{\ESS} & 2.53 & 2.64 & 3.34 & \textbf{2.51} \\
{\ESU} & 2.72 & 2.82 & 2.84 & \textbf{2.64} \\
{\IKU} & \underline{\textbf{2.31}} & \underline{2.42} & \underline{2.46} & \underline{2.36} \\
{\GRN} & \textbf{2.93} & 3.07 & 3.49 & 3.03 \\
{\ENG} & 2.53 & 2.48 & \textbf{2.47} & 2.51 \\
{\SPA} & 8.97 & 2.72 & \textbf{2.60} & 2.69 \\
\bottomrule
\end{tabular}
\caption{\label{LMresult-1} Character-level perplexity for setting 1 (all available data). V means the vocabulary size for BPE operation. Bold numbers represent the best model for each language while underlined numbers show the best model for each tokenization.} 
\end{table}

\begin{table}[t!]
\centering
\begin{tabular}{lrrrr}
\toprule
Language & Morfessor & BPE (V=500) & BPE (V=5k) & Character\\
\midrule
{\ESS} & 2.77 & 3.14 & 3.23 & \textbf{2.64}\\
{\ESU} & 2.98 & 3.74 & 3.61 & \textbf{2.89}\\
{\IKU} & \textbf{2.59} & 3.02 & 2.96 & 2.61\\
{\GRN} & 3.16 & 3.44 & 3.41 & \textbf{2.97} \\
{\ENG} & \underline{\textbf{2.40}} & \underline{2.81} & \underline{2.59} & \underline{2.56}\\
{\SPA} & \textbf{2.66} & 3.23 & 3.18 & 2.94 \\
\bottomrule
\end{tabular}
\caption{\label{LMresult-2} Character-level perplexity for setting 2 (New Testament only).}
\end{table}

Tables \ref{LMresult-1} and \ref{LMresult-2} summarize the language modelling experiment results excluding FST segmentation for the 1) \textit{all data} setting and 2) \textit{New Testament only} setting, respectively.
It is suggested that the character and Morfessor models might work better than BPE models for polysynthetic languages. 
As for the 1) \textit{all data} setting, tokenization by character resulted in the best performance in modelling {\ESS} and {\ESU} while Morfessor models performed the best for {\IKU} and {\GRN}. 
BPE models with the vocabulary size of 5k worked the best with {\ENG} and {\SPA}. 
The same trend was observed for the 2) \textit{New Testament} setting for {\ESS}, {\ESU} and {\IKU}: character models performed the best for {\ESS} and {\ESU} while Morfessor led to the lowest perplexity measure for {\IKU}. 
However, the character-level model resulted in the lowest character-level perplexity for {\GRN} while the Morfessor model was the best for {\ENG} and {\SPA} for the 2) \textit{New Testament} setting. 
While it is unclear why a certain tokenization method worked better for a language, it is speculated that BPE might not be well-suited in segmenting polysynthetic languages given their morphological richness. 
A word in a polysynthetic language might consists of several morphemes that are not immediately retrievable based on the surface form. As shown in \Cref{ilg:aghnaaguq}, a word in {\ESS} may contain a root, a derivational suffix and inflexional suffixes, which may look different in the surface form depending on the morphophonological rules that apply to the suffixation. 
For example, the derivational suffix (\texttt{-$\sim$:(ng)u}) in \cref{ilg:aghnaaguq} has two morphophonological symbols ($\sim$ and :), the latter of which applies to delete the gh ending of the root \citep[for details see][]{Jacobson:2001}. 
Given such characteristics of polysynthetic languages, the fact that character models worked the best for {\ESS} and {\ESU} might mean that those languages were hard to segment with unsupervised segmentation methods like Morfessor and BPE. 
Segmenting those languages might require getting at the underlying form with linguistically motivated segmentation rather than segmenting the surface form only. 
Even though Morfessor models worked the best for {\IKU} under both settings and for {\GRN} under the 1) \textit{all data} setting, the difference between the Morfessor models and character models are quite small. 

It should be noted that the hyperparameters for Morfessor and BPE operations are not optimized. 
While the BPE models with the two hyperparameters (V=500 and V=5k) did not result in the best model for any of the polysynthetic languages, it is possible that different hyperparameters might result in better (or worse) perplexity measures. 
In a similar note, different datasets in a language might work differently with Morfessor tokenization: the Morfessor segmentation was the best in modelling {\SPA} under the 2)\textit{ New Testament} only setting, but it was the very worst under the 1)\textit{ all data} setting.

\begin{landscape}
\begin{table}[t!]
\centering
\begin{tabular}{l|r|r|rrrr|rr|r}
\toprule
Language & Setting & Morfessor & \multicolumn{4}{|c|}{FST} & \multicolumn{2}{|c|}{BPE} & Character\\
& & & - & +BPE(V=500) & +BPE(V=5k) & +char & V=500  & V=5k & \\
\midrule
{\ESS} & All & \underline{2.53} & \textbf{2.30} & \underline{2.35} & \underline{2.36} & \underline{2.33} & \underline{2.64} & 3.34 & \underline{2.51}\\
{\ESS} & NT & 2.77 & \underline{\textbf{2.25}} & 2.41 & 5.38 & 2.42 & 3.14 & \underline{3.23} & 2.64\\
{\GRN} & All & 2.93 & 2.74 & 2.70 & 2.70 & \textbf{2.69} & 3.07 & 3.49 & 3.03 \\
{\GRN} & NT & 3.16 & 4.82 & 2.93 & \textbf{2.65} & 2.68 &  3.44 & 3.41 &  2.97\\
\bottomrule
\end{tabular}
\caption{\label{LMresult-FST} Character-level perplexity including FST segmentation}
\end{table}

\begin{table}[t!]
\centering
\begin{tabular}{l|r|r|rrrr|rr|r}
\toprule
Language & Setting & Morfessor & \multicolumn{4}{|c|}{FST} & \multicolumn{2}{|c|}{BPE} & Character\\
& & & - & +BPE(V=500) & +BPE(V=5k) & +char & V=500  & V=5k & \\
\midrule
{\ESS} & All & 1903.37 & \textbf{882.16} & 1037.10 & 1097.51 & 980.49 & 2718.52 & 18157.32 & 1790.03\\
{\ESS} & NT & 3986.77 & \textbf{739.92} & 1289.70 & 891605.81 & 1353.29 & 10993.00 &  13975.35 & 2689.33\\
{\GRN} & All & \underline{287.71} & \underline{203.99} & \underline{187.76} & 189.23 & \textbf{185.96} & \underline{372.84} & 725.33 & 348.70 \\
{\GRN} & NT & 432.68 & 3988.57 & 288.63 & \underline{\textbf{171.56}} & \underline{181.21} & 673.22 & \underline{644.65} & \underline{313.79}\\
\bottomrule
\end{tabular}
\caption{\label{LMresult-FST-word} Word-level perplexity including FST segmentation} 
\end{table}
\end{landscape}

As a way to utilize rich morphology in modelling polysynthetic languages, we trained FST-based models for {\ESS} and {\GRN}. 
\cref{LMresult-FST} summarizes the character-level perplexity values for all tokenization methods including FST segmentation only and FST segmentation with character or BPE backoff strategy for {\ESS} and {\GRN}. 
For all settings, FST-based segmentation resulted in the best model for the two languages. 
The clear difference between FST-based models and non-FST-based models suggest that the Morfessor and BPE models failed to capture the morphological information present in the data. 

The fact that the FST segmentation only worked the best for {\ESS} might suggest that the FST segmentation for the language might have been more robust than {\GRN}. 
Indeed, the FST segmentation only resulted in high perplexity in modelling {\GRN} under the 2) \textit{New Testament} setting. 
With the BPE and character backoff, {\GRN} FST models still worked the best, but it is speculated that the FST morphological segmentation alone for {\GRN} might not have been reliable or the coverage of the FST was not as good as the {\ESS} FST.

\begin{table}[t!]
\centering
\begin{tabular}{lrrrr}
\toprule
Language & Morfessor & BPE (V=500) & BPE (V=5k) & Character\\
\midrule
{\ESS} & 1903.37 & 2718.52 & 18157.32 & \textbf{1790.03} \\
{\ESU} & 2244.44 & 2969.89 & 3113.87 & \textbf{1783.40} \\
{\IKU} & \textbf{1469.02} & 2185.62 & 2503.41 & 1773.08 \\
{\GRN} & \textbf{287.71} & 372.84 & 725.33 & 348.70 \\
{\ENG} & \underline{49.37} & \underline{45.87} & \underline{\textbf{44.86}} & \underline{47.83} \\
{\SPA} & 13051.98 & 75.10 &  \textbf{62.01} & 71.97\\
\bottomrule
\end{tabular}
\caption{\label{LMresult-1-word} Word-level perplexity for setting 1 (all available data). V denotes vocabulary size for BPE operation }
\end{table}

\begin{table}[t!]
\centering
\begin{tabular}{lrrrr}
\toprule
Language & Morfessor & BPE (V=500) & BPE (V=5k) & Character\\
\midrule
{\ESS} & 3986.77 & 10993.00 & 13975.35 & \textbf{2689.34}\\
{\ESU} & 4562.23 & 26323.96 & 20053.28 & \textbf{3572.00}\\
{\IKU} & \textbf{3923.01} & 14970.43 & 12581.29 & 4231.44 \\
{\GRN} & 432.68 & 673.22 & 644.65 & \textbf{313.79} \\
{\ENG} & \underline{\textbf{39.62}} & \underline{77.30} & \underline{55.01} & \underline{52.03}\\
{\SPA} & \textbf{67.93} & 158.86 & 148.47 & 106.16 \\
\bottomrule
\end{tabular}
\caption{\label{LMresult-2-word} Word-level perplexity for setting 2 (New Testament only). V denotes vocabulary size for BPE operation}
\end{table}

After comparing different tokenization methods per language, we compared different languages to see which language is easier or harder to model.
This line of inquiry has been pursued by several recent studies \citep{cotterell-etal-2018-languages, mielke-etal-2019-kind, gerz-etal-2018-relation}, where various languages are modeled using a state-of-the-art neural language model to compare relative difficulty of modelling a language with particular linguistic features. 
It should be noted that our data per language were not parallel so the comparison has to be drawn with caution. 
However, we still attempted the comparison here as comparing our models may provide insights for future studies given that we used the same or very similar RNN language models as the previous literature and that polysynthetic languages have not been discussed in this line of inquiry.
If we compared the character-level perplexity, \cref{LMresult-1} and \cref{LMresult-2} show that {\IKU} was the easiest to model under the 1) all data setting and {\ENG} under the 2) New Testament setting. 
However, character-level perplexity may not be the right metric to use to compare different languages. 
The problem with the character-level measure is that it does not tell us much about real-life applications, where the difficulty of predicting an entire word might be more meaningful. 
More importantly, the character-level perplexity underestimates the difficulty of modelling polysynthetic languages as they tend to have longer, morphologically complex words. 
In fact, when we look at the word-level perplexity, the differences between polysynthetic languages and others become clearer. 
\cref{LMresult-1-word} and \cref{LMresult-2-word} show the word-level perplexity measures for the two experimental settings. 
When considering the difficulty of predicting the next word in the languages than the next character, {\IKU} is no longer the easiest to model under any condition. 
The word-level measure clearly shows that {\ENG}, followed by {\SPA}, was the easiest to model.
Comparisons of the word-level perplexity values suggest that {\ESS}, {\ESU} and {\IKU} are quite similarly hard to model while {\GRN} is less difficult even though it is still quite harder to model than language like {\ENG} and {\SPA}. 
This observation agrees with our previous observation about the descriptive statistics of the datasets.

Of course, it might be unrealistic to expect that a model for polysynthetic languages would result in a word-level perplexity comparable to that for {\ENG} given the linguistic difference.
Polysynthetic languages tend to have longer and diverse word forms because of their richer morphology. 
Therefore, they are likely to be harder to model than other languages.
However, comparing the character-level perplexity only may result in mistakenly arguing that {\IKU} is easier to model than {\ENG}.

While the relative performance of each tokenization method for a given language stays the same regardless, the choice of the unit for the perplexity measure should be carefully made if we are to compare different languages. 
As mentioned above, the datasets were not strictly parallel across the languages even under the 2) \textit{New Testament} setting. 
Parallel texts and different evaluation methods might facilitate comparison across languages. 
For example, \citet{mielke-etal-2019-kind} uses the average surprisal (negative log-likelihood loss) per verse when comparing languages models using data fully aligned at the verse level and also suggests a statistical method to estimate the difficulty coefficient of a language given some missing verses.
Aligning a parallel corpus of polysynthetic languages and others at the verse or sentence level may lead to a more useful comparison in future research.


\begin{figure}
    \centering
    \includegraphics[width=\textwidth]{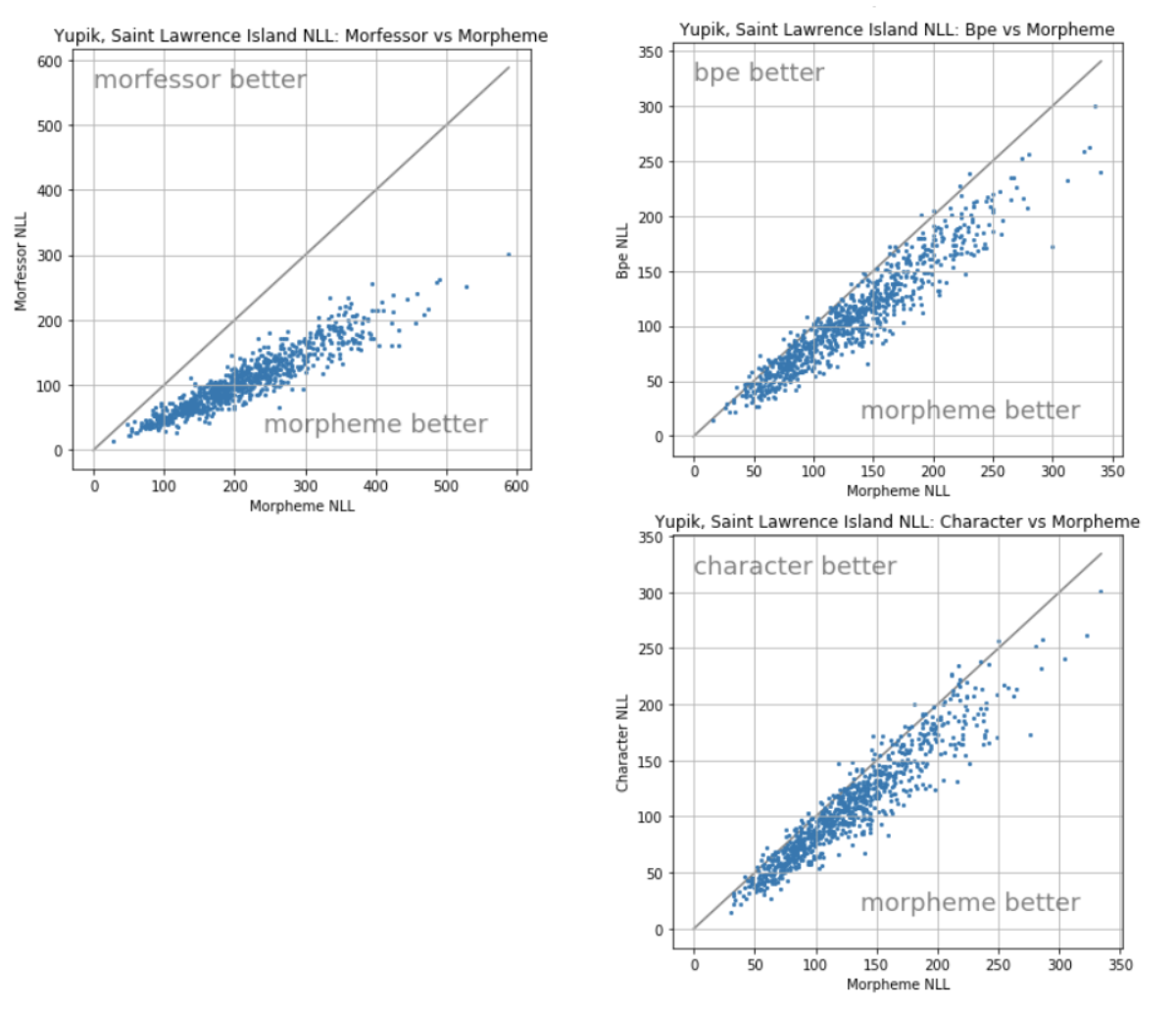}
    \caption{Model comparison in sentence-level negative log-likelihood for {\ESS}}
    \label{fig:sentence-level-loss-ess}
\end{figure}

\begin{figure}
    \centering
    \includegraphics[width=\textwidth]{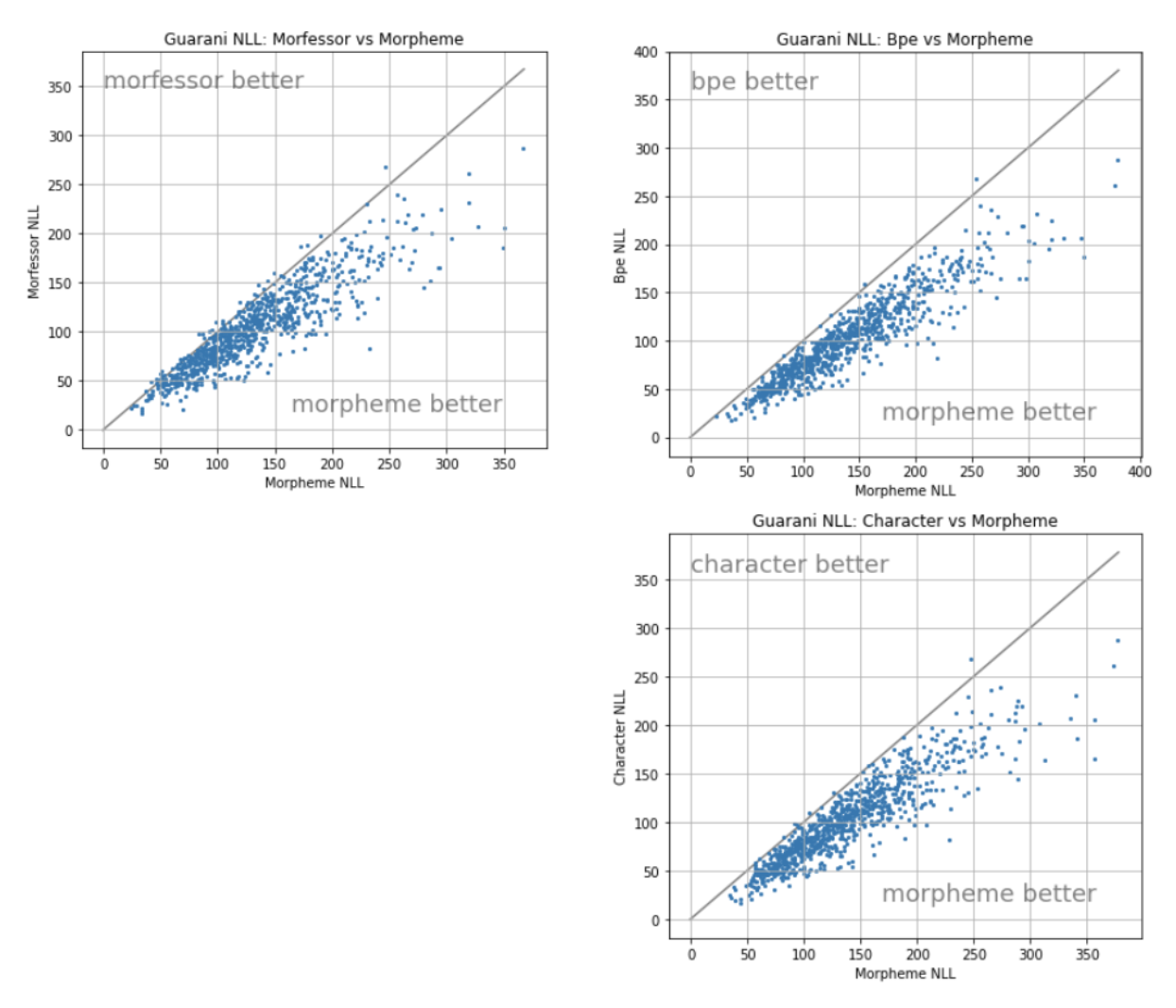}
    \caption{Model comparison in sentence-level negative log-likelihood for {\GRN}}
    \label{fig:sentence-level-loss-grn}
\end{figure}

\section{Future Direction}

The results clearly show that FST segmentation is helpful in modelling polysynthetic languages.
While we had only two languages to experiment with FST segmentation, FST segmentation with or without a backoff strategy resulted in the best model by a large margin.
\cref{fig:sentence-level-loss-ess} and \cref{fig:sentence-level-loss-grn} visualize the relative performance of the FST model v. BPE or character models at the sentence level for {\ESS} and {\GRN}, respectively. 
For both figures, points under the 45 degree line mean lower loss or better performance for the FST model than Morfessor, BPE or character model.
For both languages, it is clear that the FST models resulted in lower loss (negative log-likelihood) per sentence overall as well as the entire text.
This represents an opportunity to utilize an existing, linguistically-oriented system in aiding neural language modelling.
While FSTs might not be as helpful in modelling high-resource languages with poor morphology, they will be essential in modelling low-resource polysynthetic languages.

Another line of inquiry we are currently pursuing is comparing polysynthetic languages with other languages in terms of language modelling difficulty.
In order to compare different languages more precisely, we are using aligned Bible datasets and comparing a perplexity measure per verse. 
By modelling 149 Bibles in 94 languages, covering 24 language families, we aim to answer if polysynthetic languages are indeed harder to model than other languages and what kind of linguistic, typological features (if any) explain such difficulty.


\chapter{Applications \& Future Work\label{ch:applications}}

\section{On-device Text Prediction\label{sec:keyboard}}


One of the goals of this workshop was to make progress in providing human language technologies that can actually be used by native speakers. As smartphones become ubiquitous in native communities, text entry is becoming an increasingly important use case.

In particular, users should have access to text entry methods, namely custom keyboards, that allow them to enter text quickly and accurately. Currently, most of the languages we consider have no form of predictive keyboard available.

Our goal was to develop a pipeline for constructing custom predictive keyboards for polysynthetic languages. We wanted the keyboards to allow both automatic completion of the current unit of text being typed by the user (where units could refer to morphemes or words) and prediction of the next unit when the user input reached a boundary. Both completion and prediction rely on language models to work, so the bulk of our efforts focused on adapting trained neural network language models for on-device use.

Ultimately, we successfully built functional prototype on-device keyboards for Guaran\'i (\GRN) and St.~Lawrence Island Yupik (\ESS). To our knowledge, these would be the first open-source predictive keyboards available for these languages on the Android platform. 

\subsection{Open Source Stack\label{kb:stack}}

We chose to integrate our predictive LM models with the \texttt{android} branch of the open source Divvun toolkit\footnote{\url{https://github.com/divvun/giellakbd-android}}. Divvun was chosen since it is actively developed, and the project has a stated goal of enabling text entry for low-resource languages. The toolkit provides base IME front end source code that handles on-device keyboard display and capturing of user input. We rewrote Divvun's default back end to enable loading a trained neural LM that could be used to make future predictions based on the text buffer content the user has already typed.

\subsection{User Interface Considerations\label{kb:UIUX}}


Polysynthetic languages pose unique challenges for UI/UX design in the context of a predictive keyboard. A key question concerns the level of granularity at which predictions should be presented.

Existing keyboards almost exclusively make predictions over whole words. For polysynthetic languages, word-level prediction is problematic. For reasons introduced in Chapters~\ref{ch:intro} and~\ref{ch:background}, it isn't feasible to train an effective language model over words in languages with extremely productive morphology. Most words are composed on-the-fly, and so would not have been seen during training. Furthermore, polysynthetic morphology permits extremely long words (e.g., ``o\~nembohuguaipu'\~a'' in  Guaran\'i). The small prediction strip present on device keyboards would not be able to comfortably accommodate so many characters in a single prediction.

As a compromise, we chose to use morphemes as the unit of prediction for our keyboard prototypes. As the user types, the prediction bar presents them with either completions of the current morpheme they are in the middle of, or predictions for the next morpheme if the language model predicts they are at a morpheme boundary.

The use of morphemes as units of prediction implies that we have access to morphological analysis and segmentation tools that can generate morpheme-level training data for our language models. These tools may not be available for all languages, in which case different subword units may need to be used. One option is do modelling and prediction over BPE word chunks. However, these would likely appear unnatural to most users, since BPE segmentation is unsupervised and linguistically unaware, leading to segmentation that doesn't correspond to any natural boundaries. A better option would be to use syllables as units, since they can be extracted with a simple model that looks for consonant/vowel alternations, and do correspond to cognitively `natural' linguistic units.

\subsection{Adapting Neural Language Models for Mobile Devices\label{kb:kblm}}

\begin{figure}[t]
\centering
\includegraphics[width=0.3\textwidth]{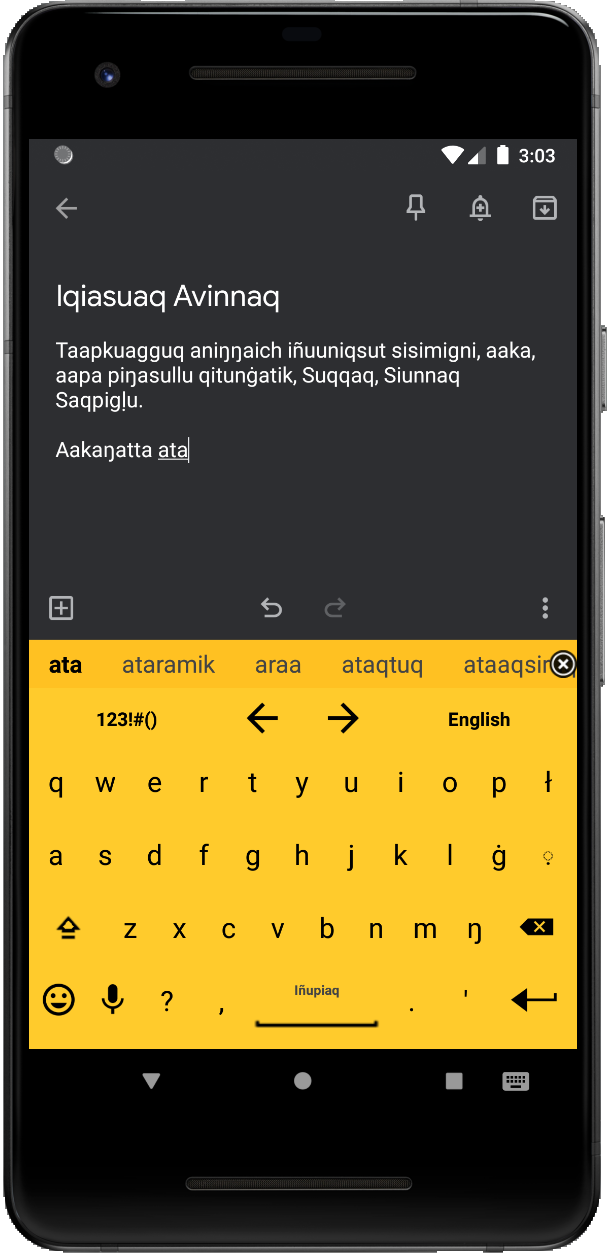} 
\caption{Sample mobile keyboard interface.}
\label{fig:phone}
\end{figure}

As shown in Figure~\ref{fig:phone}, we'd like to build an interface that uses the context typed into a buffer to present completions and predictions to the keyboard user. To do this, we need to feed the context data into a language model.

Initially, we attempted to use the SOTA PyTorch-based language models tested in Chapter~\ref{ch:lm} directly on-device. However, this proved to be technically prohibitive. First, device resources are limited, and keyboards should be lightweight --- they only account for text entry and shouldn't have a significant impact on other running applications. We set a goal of keeping keeping our model size on the order of 10Mb. Second, there is little built-in support in Android for loading and running PyTorch models. In contrast, Google provides the TensorFlow Lite(TFLite) framework for loading models trained via TensorFlow and converted for on-device use.

We attempted to convert our PyTorch models to TensorFlow using the ONNX, toolkit\footnote{\url{https://onnx.ai}} but found that the automatic converter did not support many of the operations used. Ultimately, we settled on training custom models for keyboard operation building on TensorFlow sample code.\footnote{\url{https://www.tensorflow.org/tutorials/sequences/recurrent}} We trained our models using the full desktop version of TensorFlow, and successfully exported the portion of the resulting computation graph responsible for inference to TFLite.

\begin{figure}[h]
\centering
\includegraphics[width=0.4\textwidth]{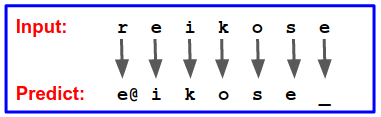} 
\caption{Language model training for keyboard.}
\label{fig:kblm}
\end{figure}

For both Guaran\'i and Yupik,
language models were trained on text from the Bible, that had been processed via the FSTs described in Chapters~\ref{ch:resources} and~\ref{ch:models} to include morpheme boundaries. The data was split as described in Chapter~\ref{ch:lm} for consistency with the language modelling experiments described there. The training data covered all available Bible
verses except the gospel of Luke (which was reserved as development data), and John (which was reserved as test data). The models were built at the character level, but with morpheme boundaries (@) marked directly on predicted symbols, as shown in Figure~\ref{fig:kblm}. This modification enabled the model to guess when a morpheme boundary was reached (i.e., a symbol with @ was predicted/typed).\footnote{Note that morpheme boundaries \emph{never} appear in the user's input buffer according to this scheme. This is different from a system based entirely on words, as the relevant boundaries, spaces and punctuation symbols, are visible.}

The model consisted of the following architecture. 
A single LSTM with 2 layers, and 200 hidden units per layer, read a 30-character context. The final hidden state of the LSTM was passed through a dense layer followed by a softmax to assign probabilities to each possible next symbol. The LSTM was trained with dropout (keep\_prob=0.75) between layers, with dropout disabled during inference. Batches of 20 contexts were used for training. Optimization was done via Adam, with initial learning rate 1.0 and learning rate decay 0.5.

When the model was loaded on the device, our custom Divvun back end sent the last 30 chars of the input buffer the user had typed through the model, and used the greedy algorithm below to generate continuations and predictions to display to the user in the keyboard's prediction bar. 

\medskip

\begin{algorithm}[H]
\SetAlgoLined
\KwResult{N prediction candidates}
    predictions = list;\
    
    \tcc{Get the LM's ranked predictions for the next char}
    
    nextFromLM = LM.predict(context[-X:];\
    
 \tcc{Loop over top N continuation points}
 
 \For{c in top N from nextFromLM}{
 
 \tcc{Greedy unroll to fill out prediction candidate}
 
 prediction = c;\
 
 tmp = (context + c)[-X:];\
 
 \While{boundary symbol (@,\textunderscore) not yet reached}
 {
  nextFromLM = LM.predict(tmp);\
  
  c = top 1 char from nextFromLM;\
  
  prediction += c;\
  
  tmp = (tmp + c)[-X:];\
  }
  predictions.append(prediction)\
 }
 \caption{Greedy continuation/prediction generation}
\end{algorithm}

\medskip

Currently, prediction stops when the model predicts a morpheme or word boundary. This stopping condition can be altered as needed to, for example, avoid stopping if the current prediction is too small (e.g., a single character) or continue predicting until the total log probability of the predicted string drops below a given threshold. Predictions can also be reached by a different, less greedy search algorithm, such as a depth first search starting at the current context. However, this has a high chance of producing many candidates with the same prefix. The method used here was chosen for its simplicity, and because it ensures candidates are diverse (no two candidates can share the same initial character). User testing might be able to determine if this bias towards diverse predictions is desirable.

\subsection{Future Development\label{kb:future}}

In Chapter~\ref{ch:lm}, we evaluate our underlying language model quality via perplexity measures. Unfortunately, we did not have access to native speakers during the workshop and so could not perform direct user testing with our prototype keyboards. 

Our ultimate goal would be to push our development back to the main Divvun project, so that it can receive ongoing support, and make it into the hands of native speakers. However, there \emph{are} a number of evaluation measures that approximate the user experience related to prediction quality. 
Top-\(n\) prediction recall measures how often the correct prediction would have been shown to the user in the keyboard's prediction strip (assuming the user was typing a fixed script). Similarly, we can measure how many keystrokes a user can save by selecting a prediction (1 touch) versus typing it out (\# touches corresponding to characters in the prediction unit).

Our prototype keyboards lack certain features that are standard on more mature offerings for languages like English. First, we assume the users touch exactly the keys they intended, and that they don't make spelling mistakes. The reality of using a touch device is that input is noisy and prone to error, with touches often sensed only in the vicinity of the intended key. A noisy channel model applied to the sequence of touch points received by the keyboard can be used to auto-correct these mistakes. 

Second, our keyboard's predictions are at the mercy of the data used to train our language models. Without a whitelist of acceptable units, or a blacklist of units that shouldn't be predicted, there is nothing preventing the model from generating offensive language. Similarly, predictions can be significantly biased towards the style of the training data. In our case, the our LMs are noticeably `evangelical,' being trained almost exclusively on text from the Bible.

\section{Speech Recognition\label{sec:speech}}

Within this section, we discuss two experiments with automatic speech recognition on polysynthetic languages: preliminary experiments with Crow (cro) word prediction and experiments with \Grn{} speech recognition. First, we describe previous work on speech recognition for polysynthetic languages as well as some of the inherent difficulties that arise when constructing speech corpora. Then, we discuss our baseline approach to end to end neural speech recognition using the Deepspeech model \citep{hannun2014deep}, the preliminary results obtained and a discussion of future directions for polysynthetic speech recognition.

\subsection{Related work}
Speech recognition for polysynthetic languages is a relatively new area of research. Much of this is due to the necessity of large transcribed speech corpora. 

\cite{klavans2018challenges} presents an overview of the challenges facing automatic speech recognition for polysynthetic languages. 
They note that there is a dearth of resources for polysynthetic languages, particularly transcribed speech corpora. These corpora require large volumes of data from skilled native language speakers.
The size of the corpora required and the linguistic, technological and language specific knowledge required make this an difficult task for communities to accomplish on their own. \cite{hasegawa2017asr} states that ``transcribing even one hour of speech may be beyond the reach of communities that lack large-scale government funding'' (as cited in \cite{klavans2018challenges}).

For Seneca, \cite{Jimerson2018} investigated the application of different ASR models to a small spoken corpus of Seneca (consisting of approximately 155 minutes of recordings). They found that GMM ASR models from the Kaldi ASR toolkit \cite{povey2011kaldi} yielded better results than neural approaches on this small dataset size -- requiring transfer learning from pretrained English ASR models and various augmentation procedures on both the text data and audio data to even approach GMM performance.

For \Grn, a relatively large speech corpus has been constructed as part of the IARPA Babel project.\footnote{Though as noted in \cite{gales2017}, the BABEL corpora are small in comparison to other corpora used in end to end neural ASR. \cite{hannun2014deep}, for example, used 5,000 hours of data.} This dataset enabled the development of several existing speech recognition systems. \cite{hartmann2016two} experimented with GMM and DNN models on several of the BABEL languages including \Grn, finding overall better performance for DNN models. Their main contribution was innovative data augmentation techniques. They sampled noise from sections of the BABEL dataset without speech data.\footnote{These sections are denoted as \textbf{<no-speech>} in the transcription files} This noise was then injected into the regular transcribed data at a signal to noise ratio between 0 and 20 db. An additional data augmentation method employed by \cite{hartmann2016two} involved speed peturbation. Previous research \cite{ko2015audio}, found that sampling the audio signal at different rates was an effective data augmentation technique. Using these two methods in combination, \cite{hartmann2016two} see a reduction in word error rate from 46.7 to 45.2.

\cite{gales2017} also worked with \Grn. They use an end to end neural approach, as we do, but they leverage stimulated network training. Stimulated network training aims to train networks where nodes with similar activation properties are grouped together \cite{gales2017}. Their paper also discusses a number of optimization methods for keyword search in speech data. They obtain a WER of 49.5 for their \Grn{} ASR system using stimulated network training. 
\subsection{Methodology}

\subsubsection{Deepspeech}
\cite{hannun2014deep} introduces the end to end neural speech recognition system used for the following experiments. This system takes in short time fourier transform (STFT) features (referred to as `spectrogram` features in the original work). These features go through three convolutional layers with ReLU activation, and then a single bidirectional RNN. Lastly, a softmax layer is used to give a probability distribution over the possible characters in the dataset. 

We borrow from this original implementation with some modifications: instead of a simple recurrent layer, we utilize gated-recurrent units, and instead of a single hidden recurrrent layer, we utilize a number of different recurrent layers. \cite{hannun2014deep} use a non-gated recurrent final layer as they were seeking to avoid computing and storing the update, input and output gates used in Long-Short-Term-Memory (LSTM) recurrent units. As a compromise between LSTMs and non-gated RNNs, we utilize Gated Recurrent Units (GRUs). Gated Recurrent Units have an update gate but no output gate, thus saving some computation in comparison to an LSTM but also allowing the neural network to be less susceptible to exploding/vanishing gradients. We also introduce more recurrent layers after the convolutional layers with significant increases in performance at the cost of increased runtime.

\subsection{Decoding}
Language models can help improve automatic speech recognition systems by imposing constraints on the possible character co-occurances. We present results for greedy decoding, where no language model is utilized and the network's predicted character sequence is not explicitly constrained. In the future, we will incorporate language models into the speech recognition system.
\subsection{Preliminary results}
Initial results for Crow word recognition and \Grn{} speech recognition are shown in the following sections.

\begin{table}[t]
    \centering
    \begin{tabular}{|c | c c | }
        \hline
        Learning rate & WER & CER\\
        \hline
        $10^-3$ & 97.07 & 87.11 \\
        $10^-4$ & 98.97 & 85.76 \\
        \hline
    \end{tabular}
    \caption{Crow speech recognition}\label{tab:crow:asr_res}
\end{table}

\subsection{Crow}
As noted in \ref{sec:resources:crow}, the data available for Crow consists only of recordings of single words and small phrases. In addition, very little monolingual text data for Crow was available. Due to the lack of long phrases, as with the \Grn{} data, and the lack of large monolingual language resources, only a single recurrent layer was used in our model, similar to the original DeepSpeech implementation. In addition, the language model created from a very small collection of Crow monolingual stories was given very little weight due to the low coverage of the model. Initial experiments at word prediction proved unsuccessful. The neural net simply produced all spaces for output.

A pretrained English model trained on the Librispeech corpus was leveraged in an attempt to get any output at all from the Crow data. This pretrained model was then adapted to the available Crow data. The results from this adapted speech recognition model are shown in Table \ref{tab:crow:asr_res}. While the results produced are very poor, the network was at least producing some output at this point.

\begin{table}[t]
    \centering
    \begin{tabular}{| c|c c | }
        \hline
        Number of GRU layers & WER & CER\\
            \hline
        1 layer & 92.98 & 52.75 \\ 
        2 layer & 87.85  & 47.90\\
        3 layer & 86.00 & 46.96\\
        4 layer & 82.08 & 44.40\\
        5 layer & 82.00 & 44.50 \\
        \hline
    \end{tabular}
    \caption{\Grn{} results using greedy decoding}
    \label{tab:guarani_asr_res}
\end{table}

\begin{table}[t]
    \centering
    \begin{tabular}{| c|c c | }
        \hline
        Number of GRU layers & WER & CER\\
            \hline
        1 layer & 92.36 & 51.89\\ 
        2 layer & 86.44 & 47.18\\
        3 layer & 83.74 & 45.49\\
        4 layer & 82.73 & 44.46\\
        5 layer & 81.80 & 44.45 \\
        \hline
    \end{tabular}
    \caption{\Grn{} results using greedy decoding and data augmentation}
    \label{tab:guarani_lm_asr_res}
\end{table}

\subsection{\Grn}
For \Grn, a number of different recurrent layers were used. Character and word error rates for the development dataset from the IARPA corpus using greedy decoding are shown in Table \ref{tab:guarani_asr_res}. Both the development and training dataset used only utterances between 1 and 15 seconds in length, thus the results shown are not directly comparable to \cite{hartmann2016two}. Future experiments will be conducted on all the data for more direct comparison. All models were trained for 50 epochs with a starting learning rate of $10^-4$ and learning rate annealing each epoch. 


\subsection{Future directions}

Moving forward, we will incorporate neural language models into the speech recognition systems. Currently, the results displayed utilize simple greedy predictors with no explicit language modelling or conventional n-gram based language models \citep{heafield2011:kenlm} for decoding. \cite{gales2017} use an RNN language model with Pashto speech recognition and found that it had a minor effect on speech recognition but helped significantly with keyword search. However, their approach seems to involve a neural language model during the decoding stage. Incorporating a neural language model into the architecture using adversarial networks could enable still lower error rates as the model


\chapter{Feature-rich Open-vocabulary Interpretable Language Model\label{ch:models}}
\chaptermark{Feature-rich Open-vocabulary Interpretable LM}


In this chapter, we present a novel general-purpose neural language modelling framework designed to be capable of handling a broad variety of typologically diverse languages, including languages whose morphology includes any or all of the following: prefixes, suffixes, infixes, circumfixes, templatic morphemes, derivational morphemes, inflectional morphemes, and clitics.
In this chapter we motivate our language modelling framework using examples drawn primarily from \Ess.
\Ess{} is a polysynthetic suffixing language in which words with 1 root, 0--3 derivational morphemes, and 1 inflectional are common, and words with up to 7 derivational morphemes have been attested \citep{deReuse:1994}.

\vspace{5mm}
\noindent
\begin{tabular}{llllll}
\ilg{ilg:QikmighhaakNeghtuq} & \multicolumn{3}{l}{Qikmighhaak} & \multicolumn{2}{l}{neghtuk} \\
& qikmigh & -ghhagh & -k & negh & -tuk \\
& dog & -small & \textsc{-Abs.Du} & to.eat & \textsc{-Intr.Ind.3Du} \\ \\
 & \multicolumn{5}{l}{`The two small dogs eat'}
\end{tabular}
\vspace{3mm}


In \Cref{ilg:QikmighhaakNeghtuq} we observe a sample two-word sentence from \Ess.
The first word \textit{qikmi\-ghhaak} is a noun composed of a noun root \textit{qikmigh}, a derivational suffix \textit{-ghhagh} that serves as a diminutive, and an inflectional suffix \textit{-k} that indicates the noun's case (absolutive) and number (dual).
The second word \textit{neghtuk} is a verb composed of a verb root \textit{negh} and an inflectional suffix \textit{-tuk} that indicates the verb's mood (indicative) and valence (intransitive), as well as the person (3rd person) and number (dual) of the verb's subject.
%
%
Note that it is common for the form in which a morpheme surfaces in a word to differ from the underlying lexical form of that morpheme.
In the morphemes' respective surface forms in this example, the final uvular fricative of \textit{qikmigh} and \textit{-ghhagh} are each dropped, the vowel of \textit{-ghhagh} is lengthened, and the final uvular fricative of \textit{negh} devoices to match the adjacent voiceless stop at the beginning of \textit{-tuk}.

\vspace{5mm}
\noindent
\begin{tabular}{llllllll}
\ilg{ilg:mangteghaghrugllangllaghyunghitunga} & \multicolumn{7}{l}{Mangteghaghrugllangllaghyunghitunga} \\
 & mangteghagh- & -ghrugllag- & -ngllagh- & -yug- & -nghite- & -tu- & -nga \\
 & house- & -big- & -build- & -want.to- & -to.not- & -\textsc{intr.ind}- & -\textsc{1sg} \\
 & \multicolumn{3}{l}{`I didn't want to make a huge house'} & \multicolumn{4}{r}{\citep[pg. 43]{Jacobson:2001}} \\
\end{tabular}
\vspace{3mm}

In Example~(\ref{ilg:mangteghaghrugllangllaghyunghitunga}), a single Yupik word represents an entire sentence.
The word consists of a noun root \textit{mangteghagh}, a derivational suffix \textit{ghrugllag} that serves as an augmentative, a verbalizing derivational suffix \textit{ngllagh}, a verb-elaborating derivational suffix \textit{yug}, another verb-elaborating derivational suffix \textit{nghite}, and inflectional suffixes \textit{tu} and \textit{nga} that mark mood (indicative) and valence (intransitive), as well as the person (1st person) and number (singular) of the verb's subject.


\section{Language Model Desiderata\label{sec:lm_desiderata}}

A language model capable of effectively modelling the full linguistic diversity found in human languages, including \Ess{} and similar endangered and polysynthetic languages, should have the following desiderata.

\subsection{Flexibility with respect to language typology}


Typical methods of categorizing languages by morphological type include isolating, fusional,
agglutinative and polysynthetic. There are also morphological affix types such as prefixes,
suffixes, circumfixes, infixes and templatic morphology, and processes such as compounding
and incorporation.  

One can think of isolating languages as those (almost) without productive morphology, such as 
Chinese and Vietnamese. These languages are well served by existing approaches to language
modelling which treat the word as the fundamental unit.

Fusional languages are those where a morpheme may represent multiple morphological or syntactic
features. Most well-known Indo-European languages are of this type. They may also have complicated,
irregular, or lexicalised phonological processes occurring when morphemes are joined together. 
Consider for example Catalan \emph{tener} `to have'---\emph{tinc} `I have'---\emph{tinga} `I have'. The 
stem is \emph{ten-}, \emph{-er} is the formant of the infinitive, \emph{-c} is the formant of the first
person singular present indicative and \emph{-nga} is the formant of the first and third person present subjective. A 
vowel change in the stem occurs when the suffixes are attached to the stem. This example has two fusional
features: multiple features per morpheme and stem-internal phonological changes caused by affixing. These
languages are fairly well dealt with in existing approaches, the number of forms that can be generated
by these processes may be larger than in isolating languages, but is essentially a finite-set.

As mentioned, current \emph{ad hoc} methods work fairly well with isolating and fusional languages, where
there are a finite number of forms for a single word. Out of vocabulary items are a problem, but are typically
related to unseen new stems rather than forms of seen stems. Agglutinating and polysynthetic languages
have this problem too, but in addition they have the problem of unseen forms of previously seen stems.

In agglutinating languages --- and in polysynthetic languages to an even greater extent --- words are 
typically made up of many morphemes concatenated together. These are typically with prefixes or suffixes,
or a combination. The Yupik example in (\ref{ilg:QikmighhaakNeghtuq}) is an example of suffixing, and indeed
Yupik is an exclusively suffixing language. Guaraní combines suffixes, which are primarily for tense, aspect, and mood (TAM) markers and 
subordination, with prefixes for valency changing and agreement. This is illustrated in Example~(\ref{ilg:guarani1})
where the \emph{ai-} prefix indicates first-person singular agreement, and the \emph{-se} suffix indicates 
volitional mood, and in Example~(\ref{ilg:guarani2}) where the \emph{ña-} prefix indicates agreement and the \emph{-va} 
suffix indicates nominalisation.

\vspace{5mm}
\noindent
\begin{tabular}{ll}
\ilg{ilg:guarani1} & Aikosénte \\
                  & Ai-ko-se-nte \\
                   & {\sc sg1}-live-{\sc vol}-{\sc just} \\
                 & `I would just like to live' \\
\end{tabular}
\vspace{3mm}

\vspace{5mm}
\noindent
\begin{tabular}{ll}\label{ex:guarani2}
\ilg{ilg:guarani2} & ñaha'arõ'ỹetéva \\
                  & ña-ha'arõ-'ỹ-ete-va \\
                   & {\sc pl1}-wait-{\sc neg}-{\sc ints}-{\sc rel} \\
                 & `that we did not expect at all' \\
\end{tabular}
\vspace{3mm}

The negative form of Guaraní verbs is formed by a circumfix of two morphemes, \emph{nd-} and \emph{-i}.
These circumfixes go around verbal derivations, agreement and (TAM) markers etc, as in (\ref{ex:guarani3}).

\vspace{5mm}
\noindent
\begin{tabular}{ll}\label{ex:guarani3}
\ilg{ilg:guarani3} & ndojuhumo’ãi \\
& nd-o-juhu-mo’ã-i \\
& {\sc neg}-{\sc 3}-find-{\sc fut}-{\sc neg} \\
\end{tabular}
\vspace{3mm}

In Chukchi the comitative case is made up of a circumfix of two morphemes,   /ɣa/- and   -/ma/. The
noun   /ławt/ `head' forms the associative singular  /ɣaławtəma/ by combining these and 
adding an epenthetic schwa.

Infixes are morphemes that break a given stem and appear inside it. For example in Seri, a language
spoken in the north-west of Mexico. It uses infixation after the first vowel in the root to 
create forms with number agreement. For
example, \emph{ic} `to plant', \emph{i}\{\emph{tí}\}\emph{c} i `did she plant it?' vs. \emph{i}\{\emph{tí}\}\{\emph{tóo}\}\emph{c} `did they plant it?'.

In languages with templatic morphology, the root is typically represented as a consonant template,
e.g. in Maltese, \emph{k-t-b} `book'. Inflection takes place by ``filling'' the slots in the root with 
other templates, such that e.g. \emph{ktieb} `book' (singular), \emph{kotba} `books', are formed
by combining the root with the vowel templates \{ø-ie, o-ø\}, and in the plural the suffix \emph{-a}.

An ideal language model would be able to encode all of these types of morphology in a generic and 
compositional manner without using language- or typology-specific tricks or assumptions (e.g. productive 
morphological processes are exclusively suffixing).\footnote{We would note that treating words 
as basic units can also be considered to be a language-specific trick designed for isolating and fusional languages.}

It should allow for arbitrary subsets of characters in a given string to form meaningful, compositional units.


\subsection{Ability to incorporate external knowledge sources as features\label{sec:incorporate_features}}
In high-resource settings, neural networks commonly function as effective feature extractors \citep{2016:Goodfellow:etal}. 
In very low-resource settings such as \Ess, extreme data sparsity means that neural models are likely to have insufficient data to effectively extract such reliable features. 
To alleviate this issue, our language model should be capable of incorporating a rich array of features from supplementary knowledge sources when insufficient data conditions prevent learning them. 

Finite-state morphological analyzers \citep{2003:Beesley:Karttunen} in particular represent a mature technology capable of serving as a reliable source of rich linguistic features.
In the Yupik \Cref{ilg:QikmighhaakNeghtuq} above, we make use of the finite-state morphological analyzer of \cite{chenschwartz:LREC:2018}.
At a minimum, we expect such an analyzer to decompose a Yupik word, providing morpheme boundary information and the associated constituent morphemes.
We expect that in most cases a morphological analyzer should also provide the underlying orthographic form of each root morpheme and each derivational morpheme, the set of linguistic features such as noun case, verb mood, person, and number associated with each inflectional morpheme, and the underlying type of each morpheme (such as noun, verb, nominalizing suffix, etc).
In the some cases, an analyzer might also provide information regarding the phonemes in each morpheme.


\subsection{Open vocabulary\label{sec:open_vocab}}

In high-resource languages, especially those that are analytic rather than synthetic, a common approach is to treat morphologically-distinct variants (such as \textit{dog} and \textit{dogs}) as completely independent word types, rather than inflected variants of a common root.
In polysynthetic languages in general, and in Yupik in particular, encountering previously unseen word forms is pervasive and should be considered the norm rather than the exception.
In very low-resource settings, it is especially important that our language model be able to robustly handle and predict out-of-vocabulary tokens.
Language models with a closed vocabulary are not viable in such settings.
Instead, we require an open vocabulary language model in which the probability of a token given a history can be robustly calculated even when that token was not present in the training data.



%
%


\subsection{Interpretability of predicted units}

By definition, a language model provides a probabilistic model over a sequence of linguistic units.
In other words, a language model must be able to provide a probability distribution over the identity of the current linguistic unit given a history representing the preceding linguistic units in the sequence.
We use the term linguistic unit to refer to an instance of any well-defined linguistic level of analysis, such as a word, a morpheme, a syllable, a phoneme, or even a grapheme.

In our language model, we require that the computational mechanism that implements the linguistic unit be interpretable.
For example, consider the case of a trained instance of our language model randomly generating a sequence of morphemes;
when the model generates a morpheme, we should be able to recover whatever rich features may be encoded therein (see \S\ref{sec:incorporate_features}), such as the underlying grapheme or phoneme sequence and the type of morpheme (root, derivational, inflectional, etc).
This should be the case regardless of whether the generated unit was present in the training data or not (see \S\ref{sec:open_vocab}).


\section{Sub-word language models\label{sec:morphLM}}


The rich morphology and phonology of Yupik and typologically similar languages results in an extreme type-token ratio. 
This fact coupled with a very small corpus size make the use of $n$-gram language models and recurrent neural language models over words highly unlikely to be effective.
\citet{schwartz-etal-2019-bootstrapping} examined the number of potential word forms word forms in \ess, and estimated approximately $1.27 \times 10^{23}$ morphotactically licensed word forms.
This number is approximately equal to current estimates of the number of stars in the observable universe.%
\footnote{\url{https://www.skyandtelescope.com/astronomy-resources/how-many-stars-are-there}}
While this estimate does not take into account restrictions imposed by semantic felicity, the polysynthetic nature of the language ensures an extremely high fraction of \textit{hapax legomenon} in Yupik texts, with \citet{schwartz-etal-2020-inuit-studies} reporting that approximately every other Yupik word token establishes a previously unseen word type.
In contrast to the astronomical number of potential Yupik word forms, the complete collection of fully digitized {\ess} texts available at the time of the 2019 JSALT workshop consisted of a corpus of slightly over 81,000 word tokens (see \Cref{ch:resources} for more details).
In lieu of word-based language models, we consider language models that utilize sub-word units.

Language models serve as an enabling technology for other downstream language technologies, including mobile text prediction.
These technologies are mature and widespread for many high-resource languages, but relatively immature and rare for polysynthetic languages.
In this section, we present several motivating use cases of sub-word language models for polysynthetic language.


\subsection{Prediction of next morpheme\label{sec:predict_next_morpheme}}

\begin{figure}[ht]
    \centering
    \includegraphics[width=\textwidth]{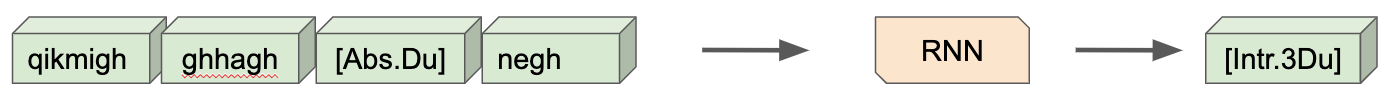}
    \caption{A recurrent neural network language model over morphemes can be used to predict the next morpheme in a sequence. In this figure, the light green boxes represent Yupik morphemes from \protect\Cref{ilg:QikmighhaakNeghtuq}, each encoded as a vector.}
    \label{fig:predict_next_morpheme}
\end{figure}

The core operation of a language model is estimating the conditional probability of a predicted next linguistic unit given a history of previous linguistic units.
\Cref{fig:predict_next_morpheme} illustrates a recurrent neural network language model that predicts the most likely next morpheme given a history of four immediately preceding morphemes, where each morpheme is encoded as a vector.
%

\subsection{Prediction of next character\label{sec:predict_next_character}}

A closely related task applicable in the context of mobile text completion is the prediction of the next character given a preceding sequence of characters.
In the polysynthetic language setting, it may be beneficial to augment such a model with a history of morphemes in situations where this information is available.  

\begin{figure}[ht]
    \centering
    \includegraphics[width=\textwidth]{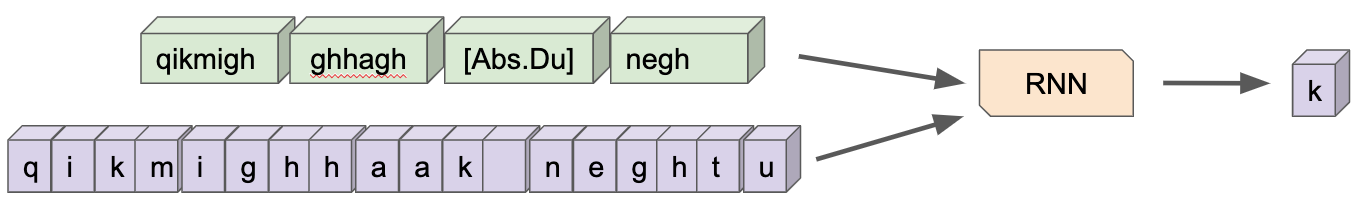}
    \caption{In a text completion setting, a more sophisticated recurrent neural network language model could predict the next character given a history of preceding characters and a history of preceding morphemes from \protect\Cref{ilg:QikmighhaakNeghtuq}. In this figure, the light green boxes represent Yupik morphemes while the light purple boxes represent characters.}
    \label{fig:tpr_char_lm}
\end{figure}


\section{Neural morphological analysis\label{sec:morph_analysis_lm}}

As discussed in \S\ref{sec:fsm}, finite-state morphological analyzers provide a mechanism for encoding linguistic knowledge in a finite-state transducer capable of analyzing a word and providing morpheme boundaries and other linguistically salient information about the underlying morphemes that comprise the word.
Recent work has explored how a finite-state morphological analyzer can be used to bootstrap a neural morphological analyzer \citep{micher:2018,schwartz-etal-2019-bootstrapping,silfverbergtyers:2019}.
Building on that work, we propose a neural morphological analyzer that directly predicts morpheme vectors, rather than predicting a sequence of strings representing an analyzed form.

\begin{figure}[ht]
    \centering
    \includegraphics[width=\textwidth]{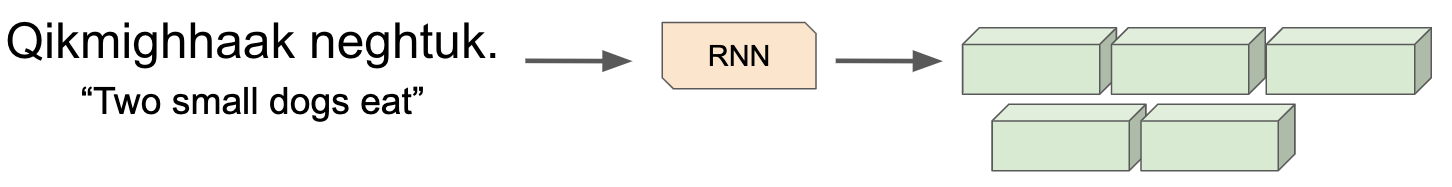}
    \caption{In a morphological analysis setting, a sequence-to-sequence model predicts a sequence of morphemes from an input sequence of Yupik characters from \protect\Cref{ilg:QikmighhaakNeghtuq}. In this figure, the light green boxes represent predicted Yupik morpheme vectors.}
    \label{fig:predict_morpheme_vectors}
\end{figure}

\section{Tensor Product Representation\label{sec:tpr}}

To satisfy the language model desiderata specified in \S\ref{sec:lm_desiderata}, we consider the Tensor Product Representation (TPR) proposed by \citet{smolensky1990}.
The use of TPRs provides a principled way of representing hierarchical symbolic information in vector spaces, such as those used as the input and output domains of neural networks.
%
%
Developing a tensor-product-based representational scheme begins by decomposing a symbolic structure into roles and fillers.  
A symbolic structure can then be represented as the \textit{bindings} of fillers to roles. 
Once decomposed, both roles and fillers are embedded into a vector space such that all roles are linearly independent from one another. 
Let $b$ be a list of ordered pairs $(i,j)$ representing filler $i$ (with embedding vector $\hat{\mathbf{f}}_i$) being bound to role $j$ (with embedding vector $\hat{\mathbf{r}}_j$). 
The \textit{tensor product representation} $\mathbf{T}$ of the information is then given by
\begin{equation}
    \mathbf{T} = \sum_{(i,j)\in b} \hat{\mathbf{f}}_i\otimes\hat{\mathbf{r}}_j \in \mathbb{R}^d \otimes \mathbb{R}^n.
\end{equation}
This TPR may itself be used as a filler and subsequently be bound to another role vector.
This process results in a TPR that represents hierarchical compositional structure.

\begin{figure}[ht]
    \centering
    \includegraphics[scale=0.25]{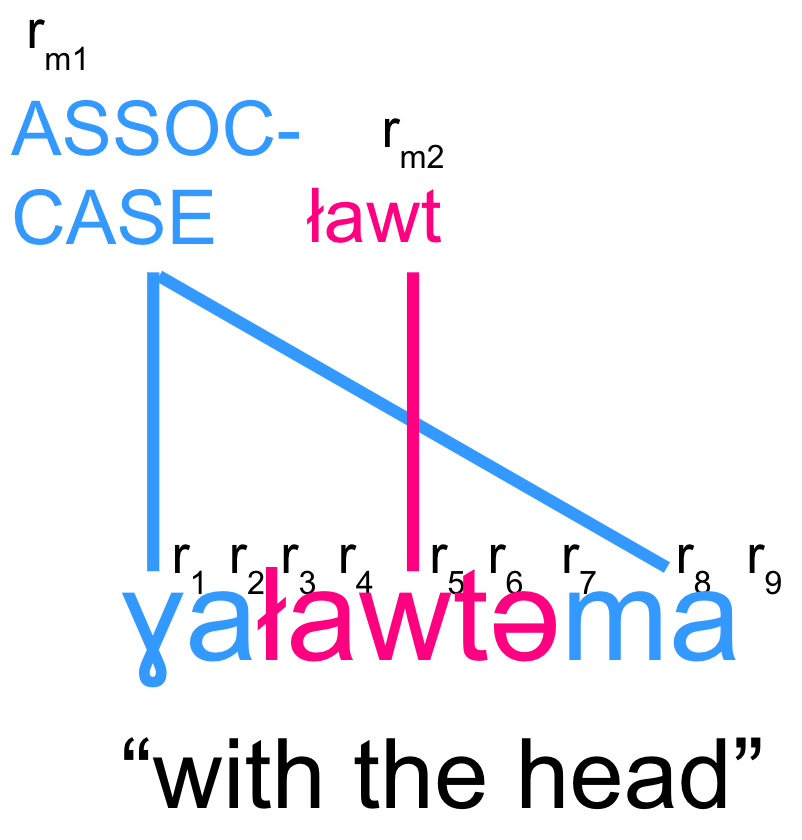}
    \caption{This sample word from {\ckt} is composed of a root morpheme \textit{\ipa ławtə} and a circumfix  \textit{\ipa ɣa{\ldots}ma}. The individual characters positions in the word comprise roles $r_1$ through $r_9$, while the characters at those respective positions comprise fillers $f_1$ through $f_9$. Roles $r_{m_1}$ and $r_{m_2}$ represent morpheme positions within the word, and are respectively filled by $f_{m_1}$ (denoting the identity of the circumfix morpheme marking associative case) and $f_{m_2}$ (denoting the identity of the root morpheme).}
    \label{fig:with_the-head}
\end{figure}


%



%
%

\subsection{Unbinding\label{sec:unbinding}}

TPRs are useful because they embed arbitrary symbolic structure in a vector space in such a way that simple linear algebra operations may be used to retrieve the form of the symbolic structure, including its compositional structure.
The core operation in retrieving this structure is called \textit{unbinding}.
We may use unbinding to query a role for its filler. Unbinding may be accomplished by any of several exact or approximate strategies. 
Exact unbinding requires linear independence of the roles; however, recent (unpublished) work points to the accuracy of approximate unbinding even in densely packed TPRs. 
In this work, we use self-addressing unbinding, as it is quick to compute and proved sufficiently accurate for our purposes.
Self-addressing unbinding retrieves the filler $\tilde{\mathbf{f}}_i$ for the role $\hat{\mathbf{r}}_i$ by simply computing the inner product between the role vector and the TPR:
\begin{equation}
    \tilde{\mathbf{f}}_i = \mathbf{T}\cdot\hat{\mathbf{r}}_i
\end{equation}
This unbinding is exact if the role vectors are orthogonal to one another. Otherwise, the intrusion of the filler of role $j$, $\hat{\mathbf{f}}_j$,
into the unbound filler of the role $i$, \(\tilde{\mathbf{f}}_i\), is given by

%

In our case, since we have a fixed filler vocabulary, we were able to snap our unbindings to the filler with the highest cosine similarity to the unbound vector with sufficient accuracy to render this intrusion irrelevant. Other unbinding strategies involve computing an inverse or pseudoinverse of a matrix of role vectors to perform a change of basis and decrease the intrusion.

%
%


\section{Morpheme vector representations from TPRs\label{sec:autoencoding}}

We use TPRs (\S\ref{sec:tpr}) to bridge the gap between the rich hierarchical symbolic information encoded in finite state morphological transducers \citep[such as][]{chenschwartz:LREC:2018} and the morpheme vectors needed by the neural models described in \S\ref{sec:morphLM} and \S\ref{sec:morph_analysis_lm}.

\subsection{Morpheme TPRs\label{sec:morpheme_tprs}}

\begin{figure}[ht]
    \centering
    \includegraphics[width=\textwidth]{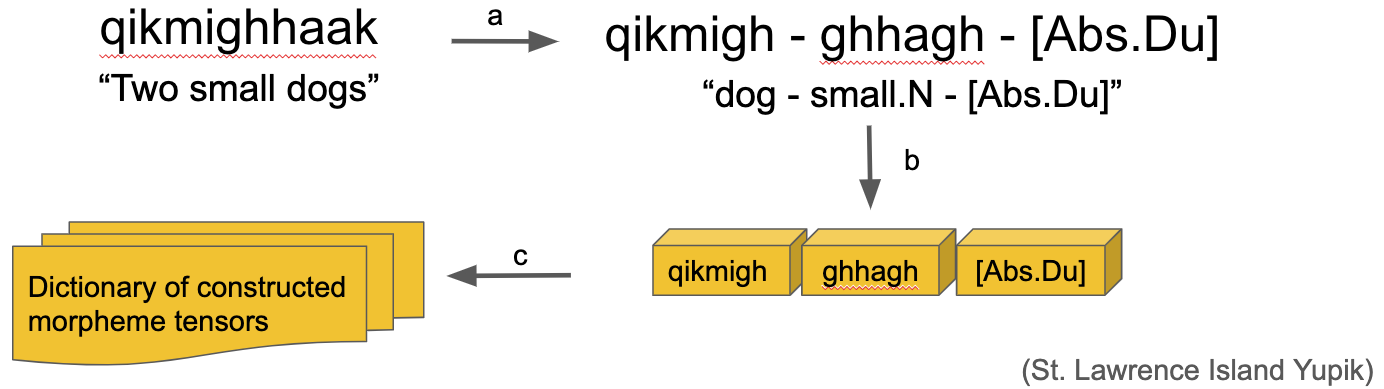}
    \caption{(a) Each word in a corpus is processed by a morphological analyzer. (b) A tensor product representation of each morpheme is calculated, resulting in one tensor per morpheme. (c) The morpheme tensors extracted from the corpus are stored in a dictionary.}
    \label{fig:morpheme_tensors}
\end{figure}

Given a language, a corpus of text in that language, and a finite-state morphological analyzer for that language, we can use the finite-state analyzer to obtain a morphological analysis for each word in the corpus.
For each morpheme provided in an analysis, we extract a collection $b$ of linguistically salient feature-value ordered pairs $(i,j)$.
Each linguistic feature $j$ serves as a TPR role; each value $i$ serves as a TPR filler.
%
%
For each such feature $j$ (such as noun case), we define $\hat{\mathbf{r}}_j$ to be a role vector representing that feature;
for each value $i$ (such as \textsc{Abs}) associated with feature $j$, we define $\hat{\mathbf{f}}_i$ to be a filler vector representing that value.
This use of TPRs enables us to jointly encode latent structural information provided by a finite state transducer with surface information in a principled manner.
%
%
This process is depicted in \Cref{fig:morpheme_tensors}.

\subsection{Learning morpheme vectors using an autoencoder\label{sec:autoencoder}}

The morpheme tensors constructed in \S\ref{sec:morpheme_tprs} are potentially very high dimensional.
Depending on how much linguistic information is encoded in each tensor, the morpheme tensors may consist of approximately $10^3$ to $10^9$ floating point values per tensor.
Tensors of this size are far too large to be directly usable as morpheme representations in the neural models described in \S\ref{sec:morphLM} and \S\ref{sec:morph_analysis_lm}.
\begin{figure}[ht]
    \centering
    \includegraphics[width=\textwidth]{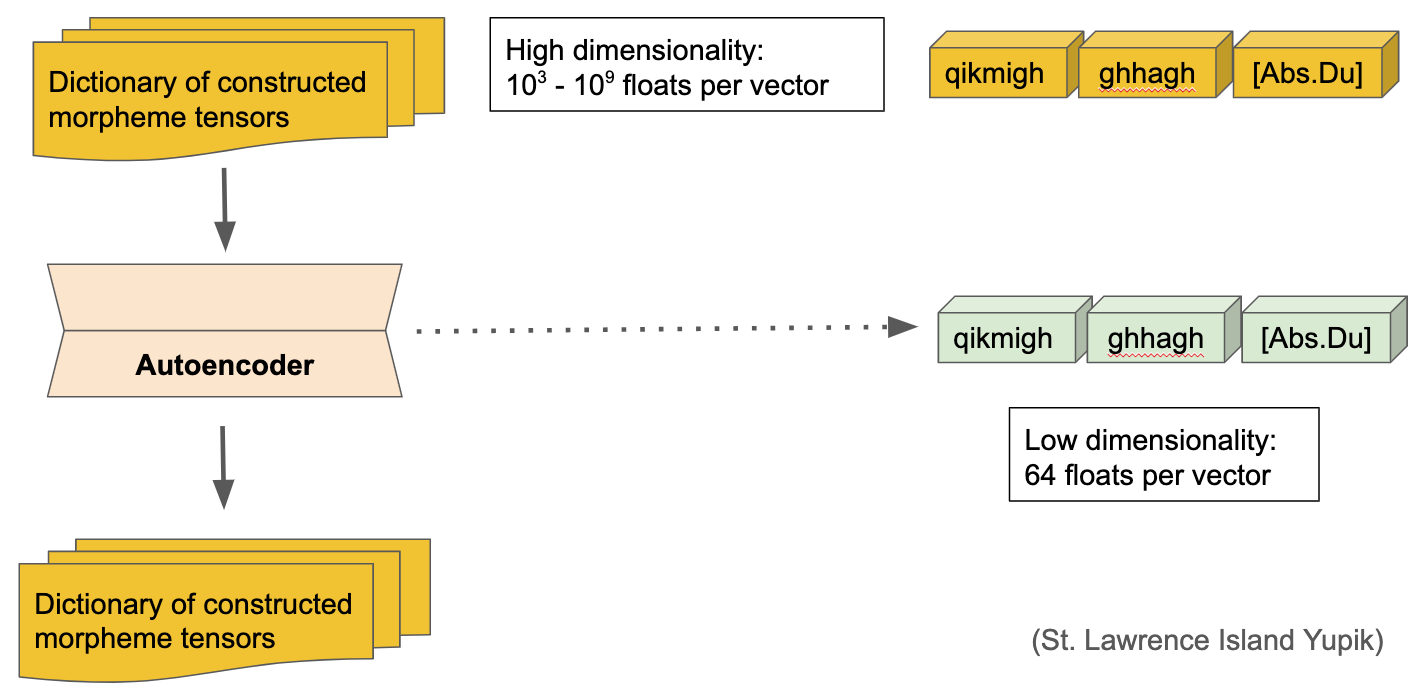}
    \caption{An autoencoder trained is on the dictionary of morpheme tensors.}
    \label{fig:morpheme_vectors}
\end{figure}
To learn lower dimensional morpheme vectors, we make use of an autoencoder.
The autoencoder is trained using the dictionary of previously constructed morpheme tensors.
The trained autoencoder can be used to encode a low-dimensional morpheme vector from a high-dimensional morpheme tensor by running the morpheme tensor through the first half of the autoencoder, and can be used to obtain a high-dimensional morpheme tensor from a morpheme vector by running the morpheme vector though the latter half of the autoencoder.


\section{Unbinding loss \label{sec:unbinding_loss}}

In order to effectively train the autoencoder in \S\ref{sec:autoencoder}, gold standard morpheme tensors must be compared against predicted morpheme tensors outputted by the autoencoder.
However, the morpheme tensors are very high dimensional.
%
%
%
In initial experiments, we used mean squared error as a loss function, but we found this was unable to converge for auto-encoding sparse TPRs. 

To enable effective training of the autoencoder, we therefore define a novel loss function that makes use of the information encoded in the TPR.
%
%
We define a loss function called \textit{unbinding loss} that examines the unbinding properties of a predicted morpheme tensor to answer the question, ``What filler is closest to the unbinding of each role in the TPR?''
%
%
%
For simplicity, we assume the use of self-addressing unbinding in this section (which we also used in the work presented here), but the computations are analogous with other unbinding strategies, relying only on a fixed role and filler vocabulary and a fixed number of bindings. 
We call the output TPR $\mathbf{T}$.

Given a predicted tensor, the first step to computing the unbinding loss is recursively unbind roles until the leaves of the structure are reached -- that is, unbind each role until the result of unbinding is a single vector (rather than a higher-dimensional tensor).
When this point is reached, we compute the cosine similarity between the result of unbinding and all the fillers in the vocabulary.
For example, assume a depth-3 structure is encoded in a TPR, where the fillers are character embeddings, the second level is left-to-right positional roles, and the highest level is morpheme identity.
If we want to see what is bound to the first position of the English \textit{cat} morpheme in $T$, we would first unbind from $\mathbf{T}$ as follows (assuming self-addressing unbinding):

\begin{equation}
\mathbf{f}_{cat, 1} = \mathbf{T}\cdot \hat{\mathbf{r}}_{cat} \cdot \hat{\mathbf{r}}_{1}
\end{equation}

We then get the vector of similarities $\hat{\textbf{s}}_{cat, 1}$ between this filler and the each of character embedding vectors in the vocabulary matrix $V$ as follows:
\begin{equation}
    \hat{\mathbf{s}}_{cat, 1} = \frac{\mathbf{f}_{cat, 1}\cdot \mathbf{V}}{||\mathbf{f}_{cat, 1}|| \mathbf{V}^{i}\mathbf{V}^{i}}
\end{equation}

where $\mathbf{V}^i\mathbf{V}^i$ denotes the column-wise vector norm of the vocabulary matrix (using Einstein summation notation).

This similarity vector can be used to define a probability distribution over possible fillers through the use of a softmax. We take the logarithm of the result of this computation to obtain log-probabilities. We call this distribution $P$.
\begin{equation}
    P = \log \Big( \frac{e^{\hat{\mathbf{s}}_{cat, 1}}}{\sum e^{\hat{\mathbf{s}}_{cat, 1}}} \Big)
\end{equation}
 We then treat each filler vocabulary word (in this case, each character) as a class, and compute the negative log-likelihood loss over this probability distribution. The resulting loss for the first character of cat being c is then
 \begin{equation}
    loss(\hat{\mathbf{s}}_{cat, 1}, c) = -\hat{\mathbf{s}}_{cat, 1, c} + \log (\sum_j  e^{\hat{\mathbf{s}}_{cat, 1, j}}).
\end{equation}

In this example, we focus on the loss for a single filler; however, as we consider tree-structured representations, the number of fillers needing to be checked is exponential with the depth of our representation. In practice, we were able to overcome this difficulty by parallelizing the independent matrix computations for the loss of all the position roles for a given morpheme, trading space for time. For more complex TPRs, a potential avenue would be to exploit the fact that most roles will be empty (and their unbindings thus a matrix of zeros) by replacing the loss computations for unbound roles with mean squared error (which need only push that part of the representation to 0). 

%
%

%
%


\chapter{Conclusions\label{ch:conclusion}}

In motivating this JSALT workshop on neural polysynthetic language modelling, we observed the following major assumptions (usually unstated) that are pervasive in most computational linguistics and natural language processing research:
\begin{itemize}
    \item If a technique works well on English, the technique is likely to be ``language agnostic'' and is likely to work well on a large variety of other languages. Various other high-resource languages such as Spanish, French, German, or Chinese are sometimes used in place of English.
    \item For any given word stem, there will be a relatively small number of morphological variants of that stem.
    \item Most or all of the morphological variants of any given word stem will appear in a sufficiently large corpus to enable learning of robust statistics.
\end{itemize}

Our work was built around explicitly challenging all of these assumptions, using a variety of polysynthetic languages and a variety of natural language tasks.
The polysynthetic languages that we chose to work with present numerous significant challenges. 
These languages are typologically very different from English and other widely-used high-resource languages.
There is pervasive use of derivational and inflectional morphology.
For most word stems, there are very large numbers of potential morphological variants, very few of which occur in any given corpus.
For all of the selected languages (with the exception of Inuktitut), the corpus sizes are very small (less than $60,000$ sentences).

\section{Contribution 1: Resources}

One contributing factor to the dearth of prior work on computational research on endangered polysynthetic languages is the lack of easily available corpus resources.
Nearly all endangered languages are very low resource.
Most CL and NLP researchers do not have the personal connections with members of endangered language communities that are often critical for obtaining data for use in research.
In preparation for this workshop, our team gathered together text and speech data from various sources for a variety of polysynthetic languages.
In cases where we have connections with indigenous community stakeholders and rights-holders, we have begun the process of discussions regarding community desires and possibilities for data distribution.
For data that we have obtained permission to distribute, we have initiated a process of public data hosting.

\section{Contribution 2: Machine Translation}

The main contributions of our machine translation work during this workshop are as follows.
With first access to the beta version 3.0 of the Nunavut Hansard \citep{v3-hansard}, we were able to provide feedback and best practices for preprocessing the dataset and shared knowledge about existing character and spelling variations in the dataset.
This work contributed to the data release and publication of \citet{v3-hansard}; that data is now being used in the Fifth Conference on
Machine Translation (WMT20) \Iku{}-\Eng{} news translation shared task.
Our work at the time constituted state-of-the-art performance on translation between \Iku{} and \Eng{}.
It has since been surpassed by \cite{v3-hansard}, and we anticipate future improvements through the WMT20 shared task.

We collected empirical evidence on several well-known but unresolved challenges, such as best practices in token segmentation for MT into and out of polysynthetic languages, as well as an examination of how to evaluate MT into polysynthetic languages.
We successfully used multilingual neural machine translation methods to improve translation quality into low-resource languages (\Ess{} and \Esu{}) using data from related languages (\Iku{}).
Notably, our ``low-resource'' languages were lower resource than much of the literature, and we produced improvements without the use of large monolingual corpora (which are unavailable for these languages and many other languages of interest).
We observed these improvements across both $n$-gram-oriented and semantic-oriented metrics.

There remain a number of open challenges in this space.
We encourage caution in interpreting the automatic quality metrics, as we do not yet have human judgments of translation quality for the languages examined; human judgements from the WMT20 shared task may prove particularly valuable.
Our initial results, using fairly conventional methods, for both multilingual and bilingual machine translation show promise, but we expect that there remains much room for improvement.

\section{Contribution 3: Language Models}


To our best knowledge, this paper represents the first attempt at modeling polysynthetic languages using a state-of-the-art RNN model and comparing their language modeling difficulty with that of other languages. We conduct language modeling experiments on four low-resource, polysynthetic languages ({\ess}, {\esu}, {\Iku}, {\Grn}) and two high-resource, morphologically poor languages ({\Eng}, {\Spa}), using four different segmentation methods: character, BPE, Morfessor and FST. By comparing the perplexity measure at the character level, we show that the FST segmentation method worked the best for polysynthetic languages when it was available. While the Morfessor segmentation method might improve language modeling performance for some polysynthetic languages, all the other segmentation method we considered---character, BPE and Morfessor---failed to capture the rich morphology of polysynthetic languages better than the FST segmentation that is based on linguistic knowledge of the languages. We also compared the perplexity measure at the word level to illustrate how significantly difficult it is to model polysynthetic languages. 

All in all, this presents an exciting starting point for a line of inquiries into modeling polysynthetic languages and utilizing the linguistic knowledge realized in FST in modeling such languages that are morphological rich and low resource. At the same time, we invite future research into linguistic characteristics that contribute to language modeling difficulty as we continue to investigate the effect of morphological complexity in our ongoing study.

\section{Contribution 4: Mobile \& Speech Applications}

As smartphones become ubiquitous in native communities, facilitating native-language communication through better technology will become an important aspect of language conservation and revitalization efforts. Building on freely available open source tools, we developed a pipeline for training neural language models that can run on-device, and loading them as a predictive back-end for on-device keyboards. This effort lead to working keyboard prototypes for Guaran\'i (\GRN) and St.~Lawrence Island Yupik (\GRN) --- the first ever input methods for these language varieties to include intelligent next-unit prediction and completion. Building the prototypes highlighted the unique requirements posed by polysynthetic languages. Their complex, productive morphology results in very long words, many of which would never appear in the training data available for language modeling, and which would be unwieldy to show to keyboard users as prediction candidates. We dealt with these problems by training character-level models that were aware of morpheme boundaries, and using morphemes rather than words as units of prediction.

The low-resource nature of most polysynthetic languages is particularly poignant for automatic speech recognition. While transfer learning can help alleviate some of the issues with data poverty, neural approaches to ASR are still not sufficient to enable usable systems.

\section{Contribution 5: Model Development}


In this workshop we proposed a novel framework for language modelling that combines knowledge representations from finite-state morphological analyzers with Tensor Product Representations \citep{smolensky1990} in order to enable successful neural language models capable of handling the full linguistic variety of typologically variant languages.
To support this framework, we also defined and implemented a novel loss function called \textit{unbinding loss} that enables gold standard morpheme tensors to be compared against predicted morpheme tensors.
We implemented a prototype TPR framework that we are continuing development of as part of ongoing future work.





\backmatter
\bibliographystyle{plainnat}
\bibliography{report.bib,MTcitations.bib}
\addcontentsline{toc}{chapter}{Bibliography}

\end{document}